\documentclass[a4paper,12pt]{article}

\usepackage[utf8]{inputenc}
\usepackage[T1]{fontenc}
\usepackage{times}
\usepackage{geometry}
\usepackage{graphicx}
\usepackage{amsmath}
\usepackage{amssymb}
\usepackage{natbib}
\usepackage{hyperref}
\usepackage{booktabs}
\usepackage{multirow}
\usepackage{array}
\usepackage{tabularx}
\usepackage{caption}
\usepackage{subcaption}
\usepackage{float}
\usepackage{setspace}
\usepackage{enumitem}

\hypersetup{
    colorlinks=true,
    linkcolor=blue,
    citecolor=blue,
    urlcolor=blue
}

\ifdefined\doubleSpacingVersion
\else
\fi

\title{\textbf{Self-Supervised Tree-level Biomass Estimation in Urban Environments From Airborne LiDAR and Optical Observations}}

\author{
Jose Bermudez$^{1,*}$, Zilong Zhong$^{1}$, Dominic Cyr$^{2}$, Camile Sothe$^{3}$, Alemu Gonsamo$^{1}$ \\
\\
$^{1}$ McMaster University, 1280 Main Street West, Hamilton, Ontario L8S 4K1, Canada \\
$^{2}$ Environment and Climate Change Canada, Montreal, Quebec, Canada \\
$^{3}$ Planet Labs PBC, 645 Harrison St., Floor 4, San Francisco, California 94107, USA \\
\\
* Correspondence: bermudej@mcmaster.ca (J.B.); ORCID: 0000-0002-4516-5787
}

\date{}
\graphicspath{{figures/}} 
\begin{document}

\maketitle

\begin{abstract}
While urban trees are a vital part of terrestrial carbon pools, their biomass remains less consistent and less spatially explicitly quantified than biomass in managed forests. Existing urban trees' carbon estimates are often based on sample inventories or coarse land-cover resolution products, which can limit their ability to resolve individual crowns and capture fine-scale heterogeneity across urban--rural landscapes. This study presents a crown-level above-ground biomass (AGB) estimation framework applied to an 810~km$^2$ urban--rural landscape in southern Ontario, Canada, integrating leaf-off airborne LiDAR (8--10~pulses~m$^{-2}$) and near-infrared RGB orthophotography (0.16--0.20~m) acquired by provincial mapping initiatives in 2018 and 2023. A dual-stream cross-attention network was trained on rule-based pseudo-labels to generate semantic marks for buildings, needleleaf trees, and deciduous trees, which were used to support crown delineation and broad functional-type assignment. Evaluation on independently annotated withheld tiles yielded global/mean scores across these three classes of 0.86 precision, 0.83 recall, and 0.84 Dice coefficient. Individual tree crowns were delineated by multiscale watershed segmentation in identified tree areas, and AGB was estimated using a power-law proxy of crown area and height calibrated against species-specific allometry \citep{Lambert2005} on 21,921 inventory trees. Across 18,713 inventory--segment matched pairs drawn from a 90,726-tree held-out test set, AGB prediction achieved $R^2 = 0.609$ when computed from inventory-derived crown geometry and $R^2 = 0.570$ under operational segmentation, isolating crown-delineation error as the dominant remaining source of AGB uncertainty. Aggregation of crown-level estimates to 30~m yielded total AGB stocks of 1.73~Tg in 2018 and 1.81~Tg in 2023 (811--850~Gg~C), local densities reaching ${\sim}140$~Mg~ha$^{-1}$ along the Niagara Escarpment, and a net carbon gain of 39~Gg~C over the five-year period. Deep-ensemble uncertainty maps identified regions of high epistemic uncertainty corresponding to land cover types underrepresented during calibration and were used to assign uncertain crowns to a pooled allometric equation. The framework operates exclusively on standard provincial sensor data without manual annotation and yields a publicly available bi-temporal crown-level AGB database for trees outside forests at a resolution relevant to municipal management.

\end{abstract}


\vspace{0.5cm}

\section{Introduction}

Trees outside forests, including urban, peri-urban, and street trees, constitute a critical yet systematically overlooked component of terrestrial carbon stocks. As urbanization accelerates globally, the spatial distribution and temporal dynamics of above-ground biomass (AGB) in these landscapes have become essential inputs for climate change mitigation, ecosystem service valuation, and equitable urban planning \citep{Zenonos2025, Liang2025, So2025}. Nevertheless, compared with managed forests, where extensive inventory systems and operational biomass models are well established, urban and peri-urban tree carbon assessments often rely on sampled-based inventories, municipal tree databases, or coarse-to-moderate resolution remote sensing products (30~m Landsat, 10~m Sentinel-2) \citep{Song2022, Liang2025}. These approaches provide valuable estimates but can be limited in their spatial completeness, consistency, and ability to resolve individual tree crowns in heterogeneous urban fabrics. Consequently, the magnitude, spatial pattern, and temporal trajectory of urban tree carbon stocks remain poorly constrained at the scales most relevant to municipal management.

High-resolution airborne remote sensing offers a partial pathway to address this gap. Airborne LiDAR provides precise three-dimensional structural information \citep{Lian2024, Liu2024}, while complementary high-resolution optical imagery captures spectral signatures that distinguish vegetation from infrastructure and needleleaf from deciduous canopies \citep{Michelini2022, Fischer2019}. The fusion of LiDAR and optical observations has been shown to substantially improve forest biomass estimation across diverse ecosystems \citep{Lian2024, Khan2025, Song2022}, and recent multi-modal deep learning approaches have reported $R^2$ values exceeding 0.80 for AGB prediction \citep{Lian2024, Liu2024, Lyu2025}. These accuracies were obtained in structurally uniform plantation or single-park settings (sub-10~km$^2$), often with terrestrial laser scanning calibration and species stratification. Comparable performance has yet to be demonstrated in heterogeneous urban landscapes spanning hundreds of square kilometers, where calibration relies solely on municipal inventory data and observations are collected under provincially mandated leaf-off conditions.

Three methodological gaps further limit the scalability of urban AGB mapping. First, many existing approaches require pixel-level manual annotation of vegetation and infrastructure for supervised training \citep{Kolanuvada2021, Weber2024}, which limits scalability to provincial coverage. Rule-based pseudo-labeling combined with neural-network refinement represents a potential alternative, yet it has mainly been demonstrated on small rural plots ($\sim$0.14~km$^2$) \citep{So2025} and has rarely been applied to the building--vegetation separation problem central to urban contexts \citep{Tuia2023}. Second, the relative contributions of automated crown delineation error and crown-based allometric model error to total AGB uncertainty remain poorly characterized \citep{Hadush2022, Jones2020, Duncanson2015}, even though crown-based proxies can perform well under ideal conditions. For example, \citet{Jucker2017} reported $R^2 = 0.87$ for crown-based biomass models compared with $R^2 = 0.92$ for diameter-based equations. These results suggest that crown-based allometry can provide a practical alternative to diameter-based biomass estimation for large-area mapping, provided that the uncertainty introduced by automated crown delineation is explicitly quantified. Third, multi-temporal AGB change products at crown scale remain rare in urban contexts and are seldom accompanied by spatially explicit uncertainty layers. Such layers are needed to distinguish areas where predictions are weakly supported by the training data from areas where model uncertainty is lower, thereby improving interpretation of AGB change and helping prioritize field validation.

This study addresses these gaps across an 810~km$^2$ urban--rural landscape in southern Ontario, Canada, using bi-temporal provincial airborne LiDAR and near-infrared RGB orthophotography acquired under leaf-off conditions in 2018 and 2023. The framework was calibrated using the Oakville municipal tree inventory, which contains 112,647 trees. Its main contributions are: (i) a pseudo-label-based crown-level AGB mapping pipeline that integrates leaf-off LiDAR and optical imagery and is validated using independently annotated withheld tiles; (ii) an explicit assessment of AGB error associated with crown geometry and automated crown segmentation, evaluated against 18,713 inventory--segment matched trees; (iii) a bi-temporal 30~m AGB density product with a deep-ensemble uncertainty layer to identify areas with limited training support; and (iv) a publicly released crown-level AGB database for trees outside forests at a spatial scale relevant to municipal management. Together, these contributions provide a prototype workflow for representing urban and peri-urban trees more consistently in regional and national carbon accounting frameworks.

\section{Methodology}

\subsection{Study Area}

\begin{figure}[t]
\centering
\includegraphics[width=\textwidth]{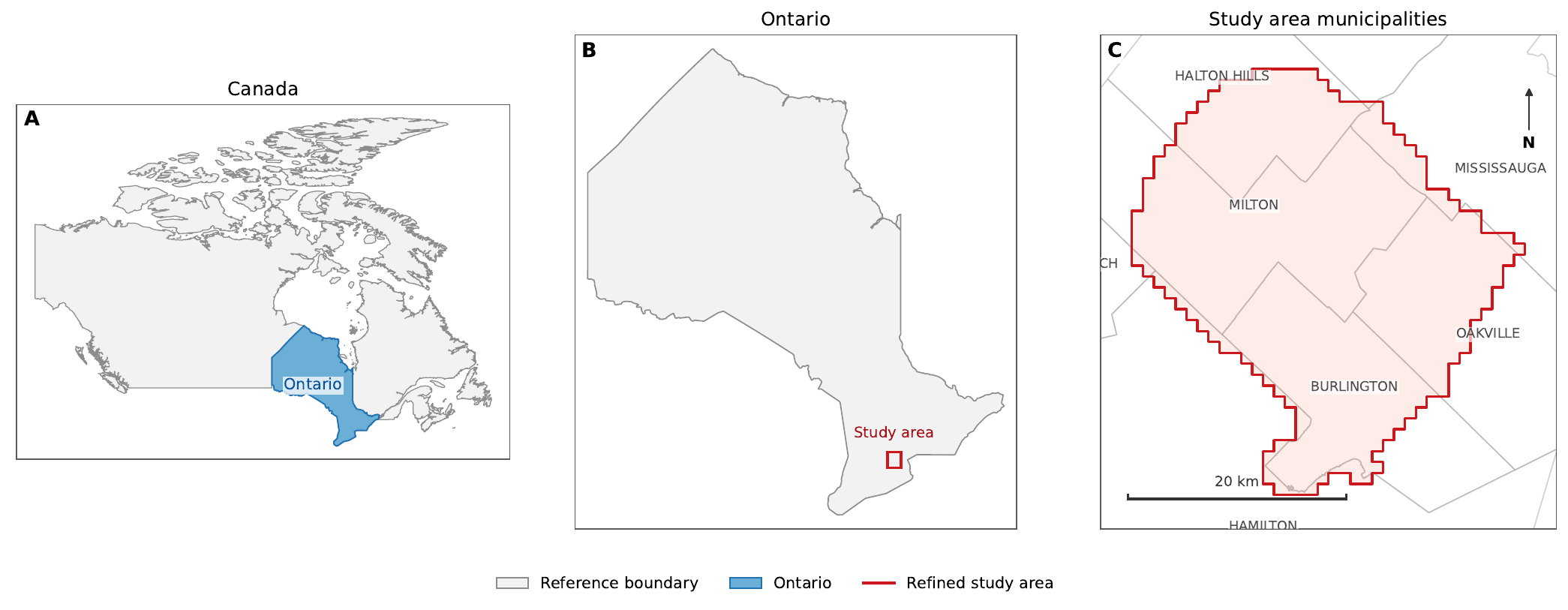}
\caption{Study area location in southern Ontario, Canada. (A) National context with Ontario highlighted. (B) Provincial context showing the study area within the Greater Toronto and Hamilton Area. (C) Municipal boundaries overlapping the study area (red outline, $\sim$810~km$^2$), defined by the availability of airborne LiDAR and optical imagery coverage across both acquisition epochs (2018 and 2023). The four Halton Region municipalities---Milton, Burlington, Oakville, and Halton Hills---account for 92\% of the study area. The Oakville municipal tree inventory provides the calibration and validation AGB reference data.}
\label{fig:study_area}
\end{figure}

The study area covers approximately 810~km$^2$ of southern Ontario, Canada, encompassing portions of eight municipalities within the Greater Toronto and Hamilton Area (Fig.~\ref{fig:study_area}). Its boundary is defined by the availability of airborne LiDAR and optical imagery coverage in both the 2018 and 2023 acquisition epochs, and it overlaps the following municipalities: Milton (279~km$^2$ overlap, 76\% of municipal area), Burlington (194~km$^2$, 72\%), Oakville (149~km$^2$, 42\%), Halton Hills (126~km$^2$, 46\%), Mississauga (31~km$^2$, 7\%), Hamilton (28~km$^2$, 2\%), Puslinch (2~km$^2$, 1\%), and Brampton ($<$1~km$^2$). The landscape grades from dense urban and suburban development in Oakville, Burlington, and Mississauga to rural agricultural land and forest patches in Milton and Halton Hills. The Niagara Escarpment traverses the western portion of the study area, supporting remnant deciduous and mixed forest stands along its slopes and associated riparian corridors. The region has a humid continental climate with warm summers and cold winters, and hosts a diverse assemblage of temperate deciduous and coniferous tree species. The Town of Oakville maintains a comprehensive georeferenced tree inventory (Section~\ref{sec:inventory}) that provides the calibration and validation reference data for this study. Wall-to-wall AGB mapping was performed across the full 810~km$^2$ study area, whereas tree-level AGB validation was restricted to the Oakville inventory footprint.

\subsection{AGB Reference Data from Oakville Tree Inventory}
\label{sec:inventory}

\begin{figure}[H]
\centering
\includegraphics[width=0.85\textwidth]{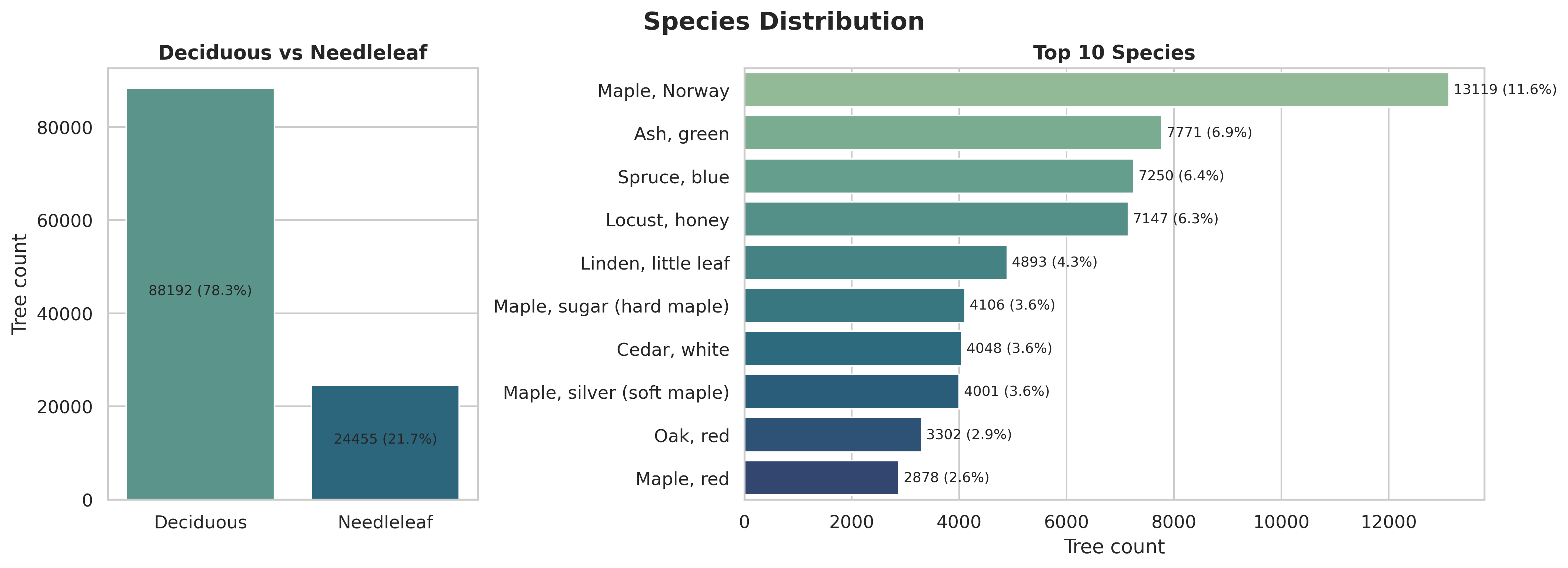}
\caption{Distributions of deciduous/needleleaf groups and the top 10 most common tree species in the Oakville municipal inventory.}
\label{fig:species_distribution}
\end{figure}

\begin{figure}[H]
\centering
\includegraphics[width=0.95\textwidth]{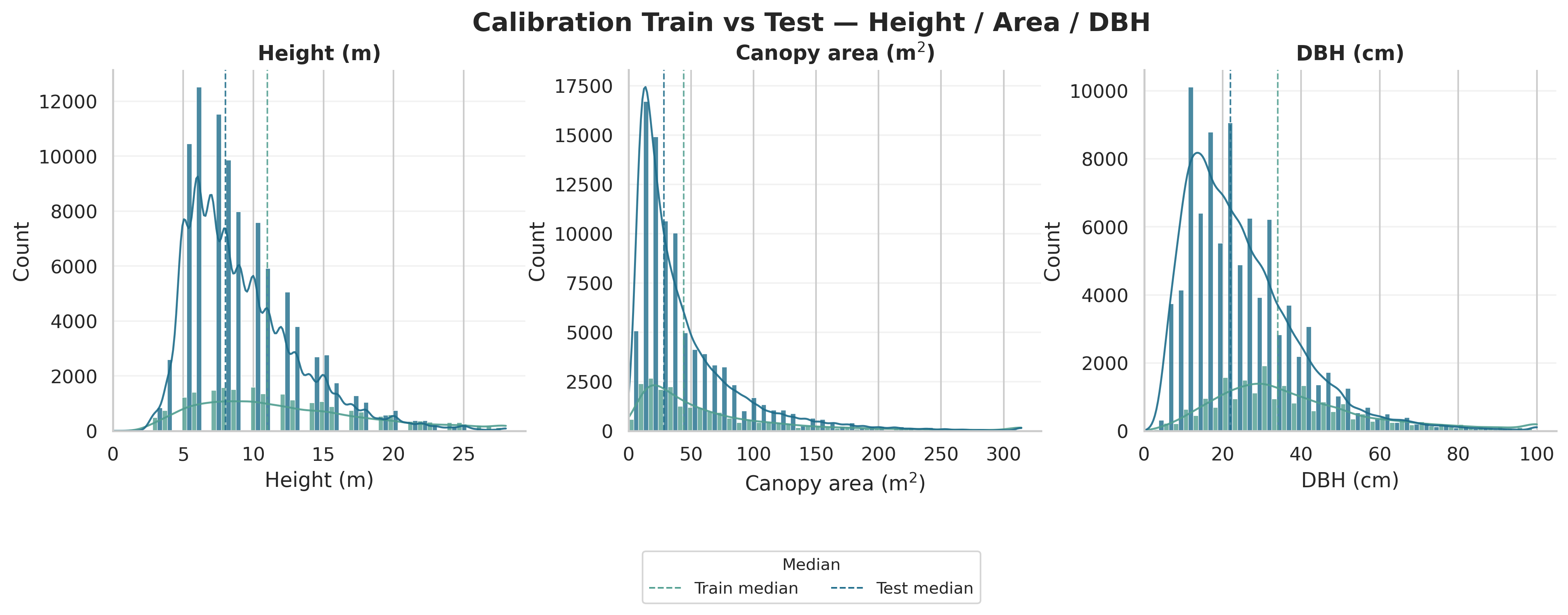}
\caption{Distribution of tree-level height (left), crown area (middle) and DBH (right) across the full Oakville municipal tree inventory ($n = 112{,}647$ trees). Overlaid dashed lines mark the training and testing medians of the calibration split (21,921 training; 90,726 held-out testing). Balanced stratified sampling enriches the calibration set with larger, rarer trees so that the allometric fit is constrained across the full size range; the training medians are therefore shifted toward larger values, whereas the held-out test set retains the inventory's natural, small-tree-dominated distribution.}
\label{fig:variable_distributions}
\end{figure}

The primary reference dataset is the municipal street-tree inventory maintained by the Town of Oakville Parks and Open Space Department \citep{OakvilleTreeInventory}, comprising 112,647 individual tree records. Each record includes species, geographic coordinates, DBH, tree height, and canopy size (crown diameter or area). Deciduous species dominate the inventory (78.3\%, 88,192 trees) relative to needleleaf species (21.7\%, 24,455 trees), reflecting the regional composition of temperate deciduous forests with coniferous plantings in urban landscapes. The most abundant species are Norway maple (\textit{Acer platanoides}; 13,119 trees; 11.6\%), green ash (\textit{Fraxinus pennsylvanica}; 7,771 trees; 6.9\%), blue spruce (\textit{Picea pungens}; 7,250 trees; 6.4\%), and honey locust (\textit{Gleditsia triacanthos}; 7,147 trees; 6.3\%) (Fig.~\ref{fig:species_distribution}). Tree measurements span a wide range: DBH from $<$5~cm to $>$80~cm, height from 2~m to over 30~m, and canopy area from $<$1~m$^2$ to $>$200~m$^2$ (Fig.~\ref{fig:variable_distributions}). The inventory covers only publicly managed trees located along streets, in parks, and on municipal properties; trees on private residential and commercial lands, which may account for a substantial fraction of total urban biomass, are excluded from both calibration and validation. The inventory was partitioned into a 21,921-tree calibration set and a 90,726-tree held-out test set using balanced stratified sampling across height, crown-area, and AGB-density strata. This design deliberately enriches the calibration set with the rarer large trees so that the allometric fit is constrained across the full size range; the held-out test set consequently retains the inventory's natural, small-tree-dominated distribution, and its size distribution is shifted toward smaller values relative to the calibration set (Fig.~\ref{fig:variable_distributions}).

\subsection{Airborne Optical and LiDAR Data and Preprocessing}
\label{sec:lidar_data}

High-resolution optical imagery was acquired through the Ontario Imagery Program, a coordinated provincial initiative managed by Geospatial Ontario (Ministry of Natural Resources and Forestry), established in 2011 on a five-year refresh cycle \citep{OntarioImageryProgram}. The program delivers four-band orthophotography (Red, Green, Blue, Near-Infrared; NirRGB) at 16--20~cm spatial resolution. Imagery is captured during early spring under cloud-free, snow-free, ice-free, and low-leaf (leaf-off) conditions to minimize canopy occlusion and to facilitate accurate ground elevation modeling. For this study, NirRGB orthophotos from the South-Central Ontario Orthophotography Project (SCOOP) acquisitions in 2018 and 2023 were used. The SCOOP data are distributed through the Ontario GeoHub (\url{https://geohub.lio.gov.on.ca/}) under the Open Government License -- Ontario. To ensure radiometric consistency between the two acquisitions, a bandwise Z-score matching transform was applied to align the 2023 imagery with the radiometric distribution of the 2018 reference (Supplementary Section~\ref{sec:supp_radnorm}).

The airborne LiDAR data were acquired through the Ontario Elevation Mapping Program (OEMP), a four-year provincial initiative running from Fall 2022 to March 2026 \citep{OntarioElevationProgram}. The point clouds were collected under leaf-off conditions (spring and fall) with a vertical accuracy of 5--10~cm and a point density of 8--10~pulses~m$^{-2}$. Owing to storage and computational constraints, the gridded elevation products were used in place of raw point clouds. Two products were obtained: (1) a Digital Terrain Model (DTM) at 0.5~m resolution, representing bare-earth elevation; and (2) a Digital Surface Model (DSM) at 0.5~m resolution, representing the top-of-canopy or building surface. The Canopy Height Model (CHM) was computed as $\text{CHM} = \text{DSM} - \text{DTM}$.

The two datasets were acquired under separate programs and processed using distinct georeferencing pipelines. The 2023 SCOOP orthophotography has a pixel size of 16~cm and a stated horizontal positional accuracy of 50~cm \citep{OntarioImageryProgram}, whereas the OEMP LiDAR conforms to the Federal Airborne LiDAR Data Acquisition Guideline \citep{NRCan2022LiDAR}, with specified horizontal and vertical errors of 35.1~cm and 10~cm, respectively. The combined worst-case horizontal offset between the two independently georeferenced products is therefore $\sqrt{0.50^2 + 0.35^2} \approx 0.61$~m. Because this nominal bound does not capture the full distribution of residual shifts encountered in practice, an empirical tile-level co-registration assessment was performed and used to design augmentation and tile-selection strategies in the processing pipeline (Section~\ref{sec:coregistration}).

\begin{figure}[!t]
	\centering
	\includegraphics[width=0.95\textwidth]{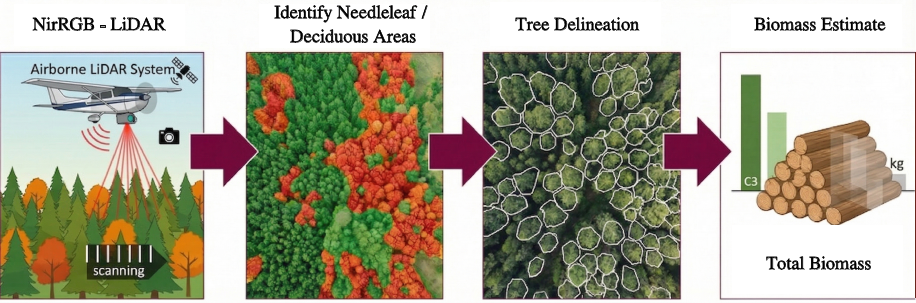}
	\caption{Four-stage pipeline for high-resolution above-ground biomass (AGB) estimation from provincial airborne LiDAR and NirRGB orthophotography. The workflow includes: (1) multi-modal data acquisition and preprocessing; (2) rule-based pseudo-label generation followed by dual-stream U-Net semantic classification to separate buildings, deciduous trees, and needleleaf trees; (3) multiscale watershed segmentation of classified tree masks to delineate individual crowns; and (4) tree-level AGB estimation from crown area and height using a calibrated allometric proxy model, with optional aggregation to 30~m AGB density.}
	\label{fig:methodology}
\end{figure}

\subsection{Processing Workflow Overview}

The proposed framework, illustrated in Fig.~\ref{fig:methodology}, integrates airborne LiDAR and optical imagery through a pseudo-label-based multi-modal fusion approach organized in four stages: (1) rule-based pseudo-label generation from LiDAR planarity analysis and spectral indices to create initial training masks for buildings, deciduous trees, and needleleaf trees without manual annotation; (2) training of a dual-stream cross-attention ConvNeXtV2 U-Net \citep{Woo2023, Vaswani2017, Ronneberger2015} on these pseudo-labels with a noise-robust Active Negative Loss (ANL-CE) \citep{ye2023active} and multitask reconstruction regularization to refine the initial masks and improve separation between built structures and tree functional types; (3) multiscale watershed segmentation of the refined tree masks to delineate individual crowns; and (4) calibration of a crown-based allometric proxy model to estimate tree-level AGB.

\subsection{Pseudo-Label Generation}
\label{sec:pseudolabels}

\begin{figure}[!t]
	\centering
	\includegraphics[width=0.95\textwidth]{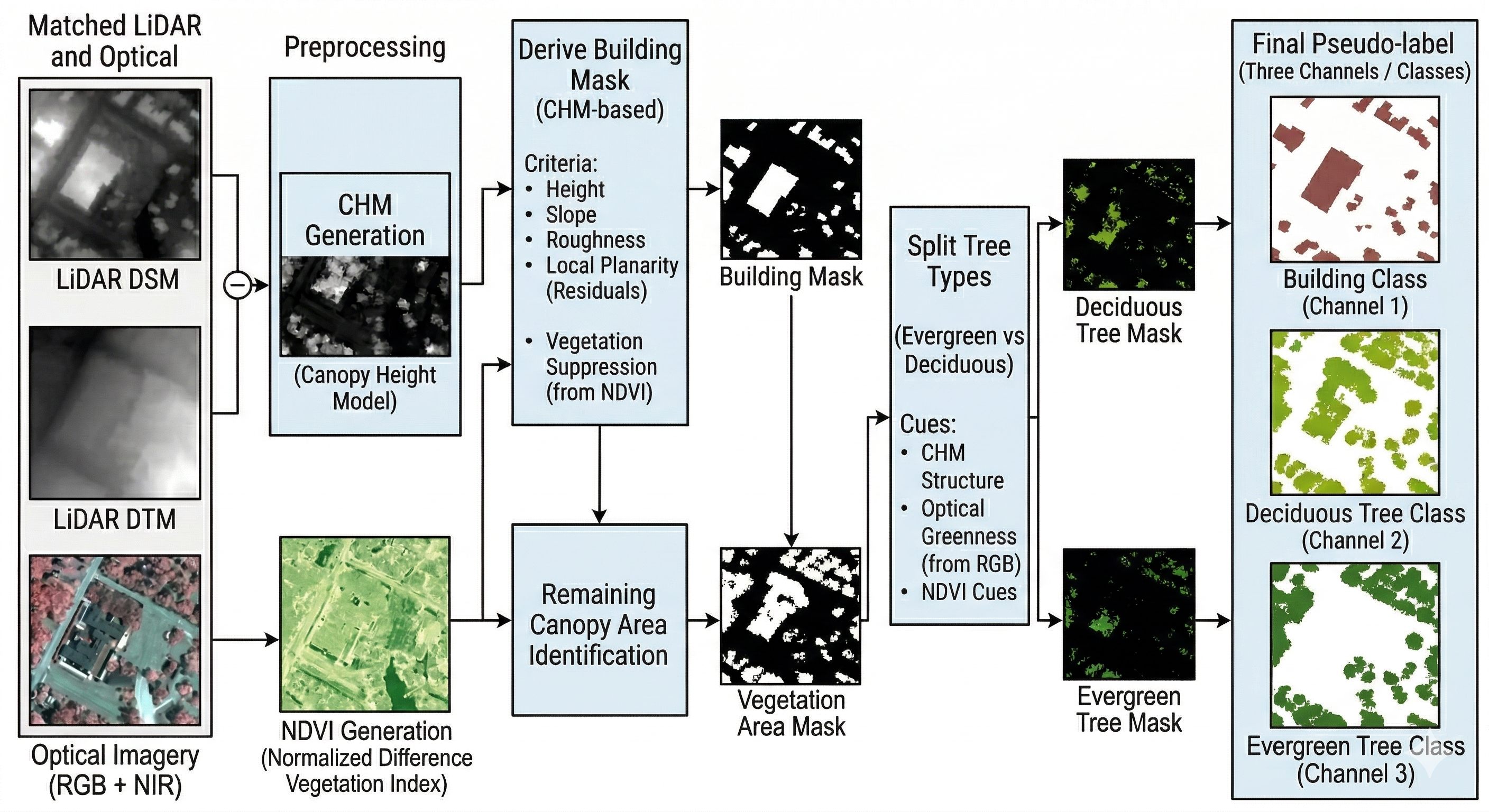}
	\caption{Pseudo-label generation framework. Rule-based classification into Buildings, Needleleaf, Deciduous, and Background provides noisy supervision for self-supervised neural network training.}
	\label{fig:pseudolabel_framework}
\end{figure}

To address the scarcity of manually labeled training data, a rule-based pseudo-label generation framework (Fig.~\ref{fig:pseudolabel_framework}) was developed that classifies each pixel into one of four categories: Buildings, Needleleaf vegetation, Deciduous vegetation, and Background. As a prerequisite, an empirical co-registration assessment was carried out to characterize the residual spatial offsets between the optical and LiDAR datasets, informing both the normalization and augmentation strategies described below.

\subsubsection{Empirical Multi-Sensor Co-registration Assessment}
\label{sec:coregistration}

Residual spatial offsets between the optical and LiDAR datasets arise from independent acquisition geometries and georeferencing errors, which can cause pseudo-label boundaries derived from LiDAR (e.g., building footprints from planarity analysis) to be misaligned with the corresponding optical image patches. To characterize this offset distribution, a tile-level normalized cross-correlation procedure (detailed in Supplementary Section~\ref{sec:supp_coregistration}) was applied to 5,356 valid tiles spanning the full preprocessing footprint across both acquisition years. Of these tiles, 661 (12.3\%) were flagged as likely misregistered, with a median offset of 1.00~m across all tiles and a 95th-percentile offset of 4.30~m among the flagged subset. To render the network robust to this range of misalignments, a random spatial translation augmentation was applied to the NirRGB input during training (Supplementary Section~\ref{sec:supp_training}), spanning the empirically observed 95th-percentile offset and forcing the network to learn features invariant to the full observed range of optical--LiDAR misregistration.

\subsubsection{Radiometric Normalization of 2023 Imagery}
\label{sec:radnorm}

To ensure consistent threshold-based classification across the 2018 and 2023 acquisitions, a bandwise Z-score matching transform was applied to align the 2023 optical imagery with the radiometric distribution of the 2018 reference (Supplementary Section~\ref{sec:supp_radnorm}). To prevent the transform statistics from being anchored to the dominant bare-soil background class in leaf-off imagery, reference statistics were computed from a visually diverse subsample of tiles selected by $k$-means clustering of MobileNetV3 visual embeddings \citep{Howard2019}, ensuring that the reference sample spans the full spectral range of vegetation and built-up classes. This normalization was applied \emph{only} during pseudo-label generation; the original imagery was retained for neural network training, in which inter-year radiometric variability acts as implicit data augmentation.

\subsubsection{Building Detection via Raster-Based Plane-Fit Residual Analysis}

Buildings were distinguished from vegetation using local planarity analysis applied directly to the CHM raster \citep{Benciolini2018, Eker2024, Ahmed2020}. For each pixel, a sliding window of size $w = 9 \times 9$~pixels was extracted, and a planar surface was fitted by least-squares regression ($z = ax + by + c$); the residual standard deviation $\sigma_r$ served as a planarity measure. Three complementary features further refined the building candidates: surface roughness (local CHM standard deviation), surface slope (Sobel gradient magnitude), and a CHM height threshold. Pixels satisfying $\sigma_r < 0.3$~m, CHM~$> 3$~m, roughness~$< 0.25$~m, and slope~$< 15^\circ$ were classified as potential buildings. The initial mask was refined by morphological opening (disk radius~$= 2$~pixels), closing, and dilation, followed by connected-component filtering (minimum footprint~$= 12$~m$^2$). Touching building instances were separated by distance-transform watershed segmentation \citep{Meyer1994}, and segments with mean NDVI~$> 0.35$ or low solidity ($< 0.55$) were rejected. The final output was a soft-weighted building confidence raster (Supplementary Fig.~\ref{fig:buildings_lidar_supp}).

\subsubsection{Needleleaf and Deciduous Vegetation Classification}

After masking buildings, vegetation pixels were classified into needleleaf and deciduous categories in three stages. \textbf{Stage 1:} pixels with CHM~$> 1.5$~m were retained as potential tree pixels \citep{Popescu2003, Eker2024} and refined by morphological opening (disk radius~$= 1$~pixel) \citep{soille1999morphological}. \textbf{Stage 2:} NDVI \citep{tucker1979red} was computed and smoothed with a Gaussian filter ($\sigma = 1$~pixel); under leaf-off conditions, needleleaf crowns retain high NDVI values whereas deciduous bare-branch crowns exhibit values near zero \citep{Cushman2023, Dietenberger2023}. Pixels satisfying both NDVI~$> 0.30$ and an HSV green color criterion (hue $\in [70^\circ,\,160^\circ]$, saturation~$> 0.15$, value~$> 0.20$) were classified as needleleaf; ambiguous pixels received fractional confidence weights $w_{\text{nl}} = \text{NDVI}^{0.5}$. \textbf{Stage 3:} tree pixels not classified as needleleaf with full confidence were labeled as deciduous, with ambiguous pixels (positive NDVI but failing the HSV criterion) receiving weights $w_{\text{dec}} = (1 - \text{NDVI})^{0.5}$ \citep{Cushman2023}.

\subsection{Self-Supervised Fusion Neural Network}

This step is designed to convert the noisy pseudo-labels into calibrated land-cover and tree-type predictions. The network architecture, loss functions, training design, segmentation validation, and ensemble uncertainty treatment are described in the following subsections.

\subsubsection{Network Architecture}

The model operates on five-channel image patches of $256 \times 256$~pixels extracted from larger processing tiles. Each patch comprises four optical bands (NIR, R, G, B) and one LiDAR-derived canopy height model (CHM) channel. The architecture is a dual-stream cross-attention ConvNeXtV2 U-Net with approximately 71.97~million trainable parameters. The optical and CHM inputs are processed by two modality-specific ConvNeXtV2 encoders \citep{Woo2023, Liu2022ConvNeXt} operating in parallel. After each encoder stage, cross-modal attention blocks \citep{Vaswani2017} enable bidirectional information exchange between the optical and LiDAR feature streams, allowing each modality to adaptively condition its representations on complementary spectral and structural information. The resulting multi-scale representations are subsequently merged by a U-Net-style decoder \citep{Ronneberger2015} with progressive upsampling and skip connections to produce full-resolution feature maps.

After decoding, the segmentation representation is refined by post-decoder modules that combine a reconstruction-guided gating signal with three complementary attention branches: spectral, local spatial, and global spatial attention. These branches are combined through a learnable fusion module, as detailed in Supplementary Section~\ref{sec:supp_refinement}, and the architecture diagram is shown in Supplementary Fig.~\ref{fig:nn_training_supp}. The fused representation is used to generate both the primary land-cover segmentation and the auxiliary tree-type prediction. The land-cover head produces three-class logits for background/other, tree, and building classes; this output provides the primary segmentation used for crown delineation and biomass mapping. The auxiliary tree-type head performs binary classification of deciduous versus needleleaf vegetation on tree pixels only, enabling the species-group-specific allometric equation selection described in Section~\ref{sec:uncertainty_allometry}. This auxiliary head is conditioned on the shared fused representation and introduces negligible additional parameters relative to the backbone. In addition to the two prediction heads, the network includes auxiliary reconstruction outputs for the optical and LiDAR inputs; these reconstructions are used exclusively during training for masked reconstruction regularization, as described in Section~\ref{sec:multitask_loss}.

\subsubsection{Multitask Learning Objective}
\label{sec:multitask_loss}

The model was trained with a multitask objective that combines land-cover segmentation, tree-type classification, and auxiliary input reconstruction. The total loss is defined as
\begin{equation}
\mathcal{L}_{\text{total}} = \lambda_{\text{LC}}\,\mathcal{L}_{\text{LC}} + \lambda_{\text{Type}}\,\mathcal{L}_{\text{Type}} + \lambda_{\text{Recon}}\,\mathcal{L}_{\text{Recon}}
\label{eq:loss_total}
\end{equation}
where $\mathcal{L}_{\text{LC}}$, $\mathcal{L}_{\text{Type}}$, and $\mathcal{L}_{\text{Recon}}$ denote the land-cover segmentation, tree-type classification, and auxiliary reconstruction losses, respectively. Full loss definitions and weighting terms are provided in the Supplementary Section~\ref{sec:supp_loss}.

The land-cover loss $\mathcal{L}_{\text{LC}}$ is computed against the rule-based pseudo-labels over the three land-cover classes (background/other, tree, and building) using an Active Negative Loss with cross-entropy as the active term (ANL-CE), a noise-robust objective designed for learning with noisy labels \citep{ye2023active}. ANL-CE combines a normalized active (cross-entropy) term that drives learning on confidently labeled pixels with a normalized passive (negative-learning) term that suppresses overfitting to label noise, weighted by $\alpha$ and $\beta$ respectively; class-volume-based pixel weighting \citep{Diakogiannis2020} is retained to mitigate class imbalance. The tree-type loss $\mathcal{L}_{\text{Type}}$ applies the same ANL-CE formulation to the binary deciduous--needleleaf classification task but is evaluated only over pixels labeled as tree; this restriction ensures that the auxiliary task focuses on vegetation type discrimination, which is required for species-group-specific allometric equation selection. The reconstruction loss $\mathcal{L}_{\text{Recon}}$ is applied to the optical and LiDAR reconstruction outputs and combines weighted mean squared error with Laplacian edge regularization and increased weighting over masked regions, following the masked reconstruction strategy of \citet{He2022}. This auxiliary objective regularizes the shared encoder by encouraging spatially coherent representations that are less sensitive to local pseudo-label noise. The reconstruction mask ratio was set to 0.6, and the modality-specific and internal weighting terms are reported in the Supplementary Section~\ref{sec:supp_loss}.

\subsubsection{Label Sources}

The framework relies on two distinct label sources, used at separate stages of the analysis. Rule-based pseudo-label rasters were generated for each tile and acquisition year and used exclusively for model optimization and fold-level checkpoint selection, as described in Section~\ref{sec:pseudolabels}. Manual annotations were used only for independent pipeline assessment after training. For this purpose, 13 tiles were manually annotated with pixel-level labels for the building, needleleaf, and deciduous classes. Each tile covered $2000 \times 2000$~pixels at 0.5~m spatial resolution, corresponding to 100~ha per tile and 1{,}300~ha in total. The segmentation metrics in Table~\ref{tab:segmentation} were computed against these manual annotations.

\subsubsection{Training Configuration}
\label{sec:training_config}

Training used bi-temporal TFRecord samples, each pairing the 2018 and 2023 $256 \times 256$ patches for a given location with their year-specific pseudo-labels. At each training iteration, one acquisition year was randomly sampled from the bi-temporal record for the forward pass, exposing the model to both acquisition years over the course of optimization.

To reduce spatial leakage, source tiles were partitioned into five non-overlapping geographic folds (Supplementary Fig.~\ref{fig:geographic_split_supp}), with all patch records from a given tile assigned to the same fold. For each held-out fold, models were fitted on patches from the remaining four folds and checkpointed on the withheld region. This tile-level split avoids the assignment of neighboring or overlapping patches from the same spatial context to both training and validation, which can produce overly optimistic performance in spatially autocorrelated landscapes \citep{Roberts2017}.

Within each fold, five experimental configurations were trained that varied in the ANL-CE passive-term weight ($\beta$), the tree-type loss weight, and the CHM-confidence weighting (Supplementary Table~\ref{tab:experiments_supp}). The five folds and five configurations together yielded a 25-member ensemble that was used for inference (Section~\ref{sec:ensemble}).

All models were trained for 75 epochs with the AdamW optimizer \citep{Loshchilov2019}, cosine learning-rate decay, an initial learning rate of $2 \times 10^{-4}$, and a 5-epoch linear warm-up. Standard spatial and radiometric augmentations, including flips, rotations, and brightness--contrast jitter, were applied during training. A random $\pm 8$-pixel NirRGB translation augmentation accounted for the empirically observed range of optical--LiDAR misregistration described in Section~\ref{sec:coregistration}. The noise-robust ANL-CE objective (Section~\ref{sec:multitask_loss}) limits overfitting to pseudo-label artifacts during this supervised training.Full hyperparameter details are provided in the Supplementary Section~\ref{sec:supp_training}.

Training required approximately 18~h per model, totaling around 450~GPU-hours for the full ensemble, with fold-level jobs parallelized on high-performance computing infrastructure.

\subsubsection{Ensemble Inference and Uncertainty Quantification}
\label{sec:ensemble}

At inference time, all 25 trained models were applied to each input tile, and their per-pixel class probability distributions were averaged to obtain the final ensemble prediction. This deep ensemble formulation \citep{Lakshminarayanan2017} improves predictive calibration relative to single-model uncertainty estimates \citep{Ovadia2019, Gustafsson2020} and provides per-pixel uncertainty maps for downstream crown-level allometric equation selection.

Inference was performed on CPU using 8 cores and required approximately 7.5~min per tile. The study area covers approximately 810~km$^2$ and comprises 936 tiles per acquisition year; processing both epochs required approximately 234~CPU-hours in total and was parallelizable across compute nodes.

Predictive uncertainty from the $M = 25$ ensemble was decomposed into aleatoric and epistemic components following \citet{Kendall2017} and \citet{Depeweg2018}. Aleatoric uncertainty represents irreducible ambiguity in the input data, such as mixed pixels, class boundaries, and crown overlap zones, whereas epistemic uncertainty captures disagreement among ensemble members and can indicate regions with limited training support. Formal definitions are provided in Supplementary Section~\ref{sec:supp_uncertainty}. Ensemble diversity was introduced through two complementary mechanisms: spatial fold diversity, since each fold was trained on a geographically distinct subset, and configuration diversity across the five experimental variants described in Supplementary Section~\ref{sec:supp_training}. For allometric equation selection, the per-pixel uncertainty maps were aggregated to the crown-segment level, producing the crown-level uncertain-pixel fraction $f_{\text{uncertain}}$ (Supplementary Equation~\ref{eq:crown_uncertainty}). Crowns for which more than 50\% of pixels exceeded a normalized uncertainty threshold of $\tau_U = 0.5$ were classified as uncertain and routed to the pooled all-trees allometric equation described in Section~\ref{sec:uncertainty_allometry}.

\subsection{Individual Tree Crown Delineation}
\label{sec:itc}

Following semantic segmentation, individual tree crowns were delineated by multiscale watershed segmentation \citep{Meyer1994} applied to the CHM (Supplementary Fig.~\ref{fig:watershed_supp}). Single-scale watershed tends to over-segment large trees and under-segment small trees in heterogeneous or overlapping canopies \citep{Berra2020, Naveed2019}; the multiscale approach addresses this limitation by applying watershed at three scales with scale-dependent Gaussian smoothing and h-maxima thresholds \citep{soille1999morphological, Naveed2019, Eker2024}: Scale 1 (small trees, $\sigma = 0.5$~m, $h = 1$~m, area $< 10$~m$^2$), Scale 2 (medium, $\sigma = 1.5$~m, $h = 2$~m, area $\in [10,\,50]$~m$^2$), and Scale 3 (large, $\sigma = 3.0$~m, $h = 4$~m, area $> 50$~m$^2$). Segments with area $< 1$~m$^2$ or height $< 1$~m were discarded; the final crown map retained segments from all three scales, with priority given to finer scales. Crown diameter was derived from crown area under the assumption of a circular crown shape \citep{Popescu2003}. Each crown was additionally screened for geometric consistency: crowns classified as needleleaf by ensemble majority vote but exhibiting crown dimensions outside the empirical range of needleleaf species in the Oakville inventory were flagged as geometrically implausible and routed to the pooled equation (Section~\ref{sec:uncertainty_allometry}).

\subsection{Biomass Estimation}
\label{sec:biomass_estimation}

Reference AGB values were computed from inventory DBH and height using the species- or genus-specific Lambert allometric equations \citep{Lambert2005}, $\text{AGB} = \beta_1 \cdot \text{DBH}^{\beta_2} \cdot H^{\beta_3}$, with functional-group (deciduous/needleleaf) coefficients applied to species not explicitly covered. Because DBH is not directly observable from airborne remote sensing, an allometric proxy model relating crown area (CA) and height ($H$) to AGB was calibrated by nonlinear least squares (NLLS) on the original AGB scale, yielding the power-law prediction
\begin{equation}
\widehat{\text{AGB}} = e^{\alpha} \cdot (\text{CA} \cdot H)^{\beta}.
\label{eq:proxy}
\end{equation}
NLLS on the original AGB scale was preferred over ordinary least squares on log-transformed data because it estimates the conditional mean directly, avoiding the retransformation bias that otherwise requires a Duan smear correction \citep{Duan1983, Baskerville1972}, and because it penalizes large-tree errors in proportion to their absolute magnitude. Three models were fitted---pooled all-trees, deciduous, and needleleaf---on a stratified training set of 21,921 trees drawn from the 112,647-tree Oakville inventory after applying biological and structural plausibility filters (DBH $\in (2,\,150]$~cm; height $\in (1.5,\,40]$~m; crown area $\leq 1{,}000$~m$^2$; AGB density $< 40$~kg~m$^{-2}$). The remaining 90,726 trees formed the held-out test set. The Oakville inventory is dominated by deciduous trees (88,192 trees, mean DBH $18.4 \pm 12.6$~cm, mean height $9.2 \pm 4.8$~m) relative to needleleaf trees (24,455 trees, mean DBH $15.7 \pm 10.3$~cm, mean height $7.8 \pm 3.9$~m). Fitted parameters are reported in Table~\ref{tab:proxy_params}.

\begin{table}[!t]
\centering
\caption{Fitted parameters of the power-law allometric proxy model $\widehat{\text{AGB}} = e^{\alpha} \cdot (\text{CA} \cdot H)^{\beta}$ (Equation~\ref{eq:proxy}), fitted by NLLS on the original AGB scale. $n_{\text{train}}$ is the training sample size.}
\label{tab:proxy_params}
\begin{tabular}{lccc}
\toprule
\textbf{Group} & $\boldsymbol{\alpha}$ & $\boldsymbol{\beta}$ & $\boldsymbol{n_{\text{train}}}$ \\
\midrule
All trees & 3.0863 & 0.8127 & 21,921 \\
Deciduous  & 4.5650 & 0.7742 & 13,359 \\
Needleleaf  & 4.3559 & 0.7391 &  8,562 \\
\bottomrule
\end{tabular}
\end{table}

\textbf{Allometric equation selection.} A crown is assigned the group-specific equation (deciduous or needleleaf) only when (i) its ensemble uncertainty fraction is $\leq 0.5$ \emph{and} (ii) if classified as needleleaf, its crown geometry lies within the needleleaf calibration domain. All other crowns receive the pooled all-trees equation. Applying a species-group equation outside its calibration domain---for example, when a large-crowned deciduous tree is misclassified as needleleaf, placing the crown volume up to 22$\times$ beyond the needleleaf calibration range---would produce unreliable extrapolations \citep{Jucker2017, Aguilar2019}.
\label{sec:uncertainty_allometry}

\textbf{Validation matching: \textit{inv} vs.\ \textit{seg} scenarios.} The 90,726 held-out test trees were spatially intersected with the final production crown segments to produce 18,713 matched tree--segment pairs (14,414 deciduous and 4,299 needleleaf trees) after applying quality filters (height ratio $0.60 \leq H_{\text{seg}} / H_{\text{inv}} \leq 1.40$; AGB density $< 40$~kg~m$^{-2}$). Two AGB predictions were computed per tree: \textit{inv}, using inventory crown geometry as an upper bound on proxy accuracy in the absence of segmentation error, and \textit{seg}, using watershed-delineated segments to represent operational accuracy. The difference between the two scenarios isolates the contribution of segmentation error from that of the allometric model.

\subsection{Wall-to-Wall Biomass Mapping and Temporal Change Analysis}
\label{sec:wall_to_wall}

The calibrated proxy models were applied to all delineated crowns to generate crown-level AGB estimates for 2018 and 2023. These crown-level vector products constitute the native high-resolution outputs of the workflow and retain individual segment geometry, height, crown area, predicted tree type, uncertainty, and AGB. For regional analysis, crown-level AGB was additionally aggregated to 30~m grid cells by summing the AGB of all crown segments whose centroids fell within each cell, producing raster AGB density (AGBD) maps in Mg~ha$^{-1}$. This aggregation converts the irregular individual-crown estimates into an area-normalized wall-to-wall product compatible with widely used 30~m Earth-observation and biomass products \citep{Dubayah2020, Song2022}, reduces sensitivity to crown-boundary and geolocation errors, and provides a practical cartographic scale for displaying the full $\sim$810~km$^2$ study area in manuscript figures. The high-resolution deliverable therefore remains crown-level, whereas the 30~m grid is used for manuscript maps, summary statistics, and temporal differencing. Temporal AGB change was computed as the pixel-wise difference between the 2023 and 2018 30~m AGBD maps. A formal minimum detectable change (MDC) analysis---propagating per-pixel classification uncertainty through the allometric model via $\text{MDC} = \sqrt{\sigma^2_{2018} + \sigma^2_{2023}}$---is deferred to future work; the temporal change results should therefore be interpreted as gross raster-level differences rather than statistically thresholded change estimates.

\section{Results}

\subsection{Segmentation Accuracy}
\label{sec:results_seg}

\begin{table}[H]
	\centering
	\caption{Segmentation performance metrics comparing rule-based pseudo-labels to ensemble model predictions, both evaluated against manually annotated tiles. Metrics are reported per class and averaged across classes.}
	\label{tab:segmentation}
	\begin{tabular}{lcccccccc}
		\toprule
		\multirow{2}{*}{\textbf{Class}} & \multicolumn{4}{c}{\textbf{Pseudo-Labels}} & \multicolumn{4}{c}{\textbf{Ensemble model}} \\
		\cmidrule(lr){2-5} \cmidrule(lr){6-9}
		& \textbf{Prec.} & \textbf{Recall} & \textbf{IoU} & \textbf{Dice} & \textbf{Prec.} & \textbf{Recall} & \textbf{IoU} & \textbf{Dice} \\
		\midrule
		Deciduous  & 0.70 & 0.89 & 0.65 & 0.79 & 0.75 & 0.90 & 0.69 & 0.82 \\
		Needleleaf  & 0.80 & 0.61 & 0.53 & 0.69 & 0.84 & 0.66 & 0.58 & 0.74 \\
		Buildings  & 0.98 & 0.84 & 0.82 & 0.90 & 0.99 & 0.92 & 0.91 & 0.95 \\
		\midrule
		\textbf{Average} & \textbf{0.83} & \textbf{0.78} & \textbf{0.67} & \textbf{0.79} & \textbf{0.86} & \textbf{0.83} & \textbf{0.73} & \textbf{0.84} \\
		\bottomrule
	\end{tabular}
\end{table}

The model is trained on the rule-based pseudo-labels and evaluated against independent manually annotated tiles (Table~\ref{tab:segmentation}). The ensemble improved on the pseudo-labels for every class and metric, raising the mean Dice coefficient by 6.3\% and the mean IoU by 9.0\%. Buildings were detected most reliably (model Dice 0.95, recall 0.92, IoU 0.91), with the model increasing building recall by 9.5\% and IoU by 11.0\% relative to the pseudo-labels. Deciduous vegetation combined high recall (0.90) with a 7.1\% precision increase. Needleleaf vegetation was the most difficult class to segment, with the lowest recall (0.66) and Dice (0.74) of the three classes, although the model still increased its recall by 8.2\% and precision by 5.0\% over the pseudo-labels, indicating reduced false positives alongside modestly improved detection sensitivity. Overall, the ensemble achieved a mean precision of 0.86, a recall of 0.83, and a Dice coefficient of 0.84 across the three classes.

\begin{figure}[!t]
	\centering
	\includegraphics[width=1.0\textwidth]{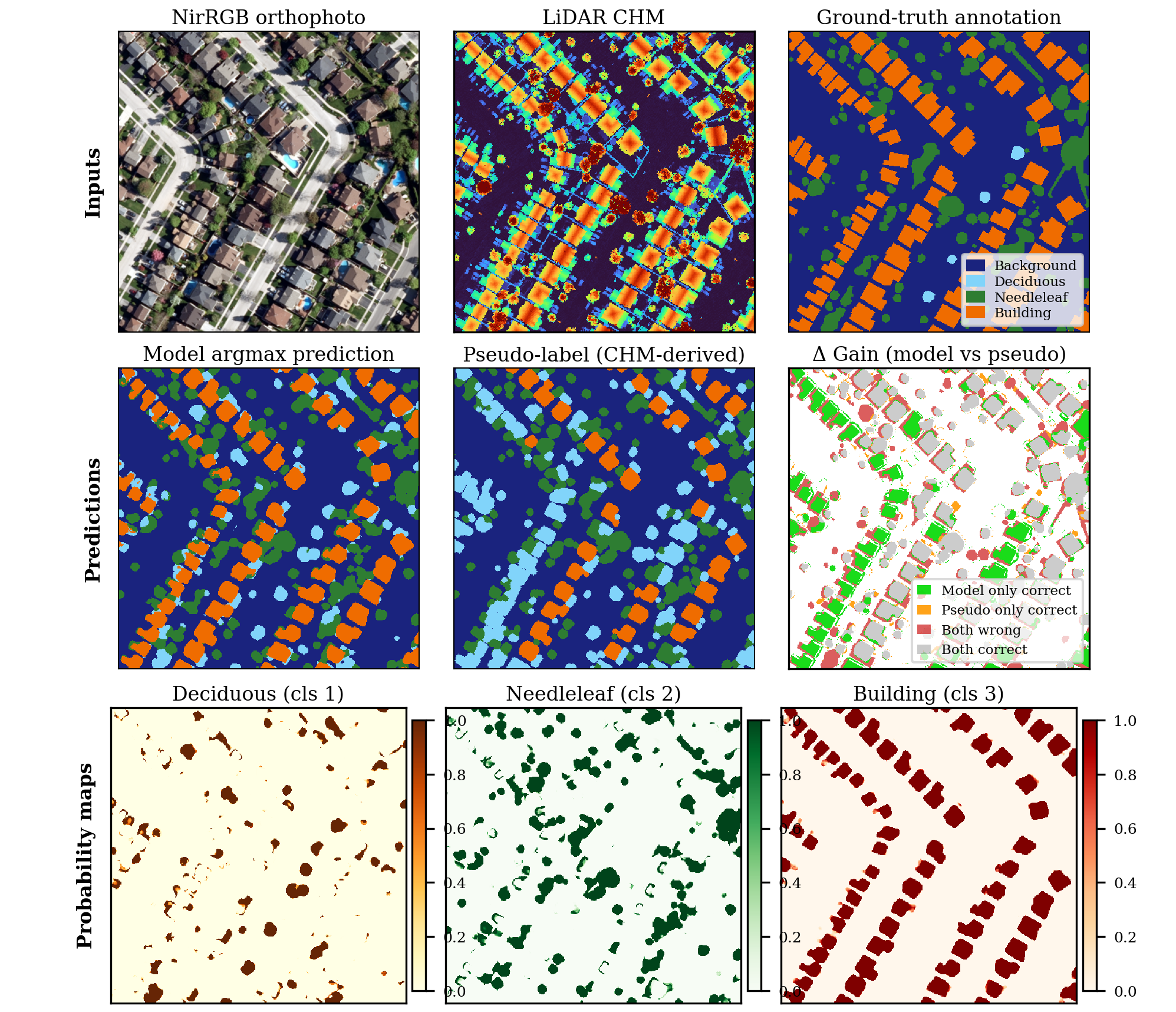}
	\caption{Representative model inference outputs for one manually annotated validation tile. \textbf{Top row:} Input NirRGB orthophotography (left), input LiDAR-derived CHM (center), and independent manual annotation (right; dark blue = background/other, light blue = deciduous, green = needleleaf, orange = buildings). \textbf{Middle row:} Final model argmax prediction (left), CHM-derived pseudo-label used for training (center), and a pixel-wise correctness comparison of the two against the manual annotation (right; green = model correct where the pseudo-label is wrong, orange = pseudo-label correct where the model is wrong, red = both correct, gray = both wrong). \textbf{Bottom row:} Softmax probability maps for deciduous vegetation (class~1, left), needleleaf vegetation (class~2, center), and buildings (class~3, right); brighter colors indicate higher class probability.}
	\label{fig:train_test_split}
\end{figure}

Fig.~\ref{fig:train_test_split} provides a qualitative illustration of the same validation comparison on one manually annotated tile. The top row shows the two input modalities (NirRGB orthophotography and LiDAR-derived CHM) alongside the independent manual annotation used only for evaluation. The middle row places the final model argmax prediction and the CHM-derived pseudo-label used during training next to a pixel-wise comparison of their correctness against the manual annotation: green marks pixels the model classifies correctly where the pseudo-label is wrong, orange marks the reverse case, red marks pixels both classify correctly, and gray marks pixels both miss. Green regions clearly outnumber orange, mirroring at the pixel level the model's net improvement over the pseudo-labels reported in Table~\ref{tab:segmentation}. The bottom row presents class-specific softmax probabilities for deciduous vegetation, needleleaf vegetation, and buildings; high-probability regions concentrate on tree crowns and roof structures, whereas lower-probability responses occur along mixed pixels, crown boundaries, and narrow vegetation--building interfaces. The building probabilities are the crispest and most confident, consistent with the high building accuracy (Dice 0.95), whereas the needleleaf responses are the sparsest and weakest, consistent with needleleaf being the hardest class (recall 0.66). Together, the panels show that the network preserves the broad pseudo-label structure while correcting local errors relative to the manual reference.

\begin{figure}[H]
	\centering
	\includegraphics[width=0.95\textwidth]{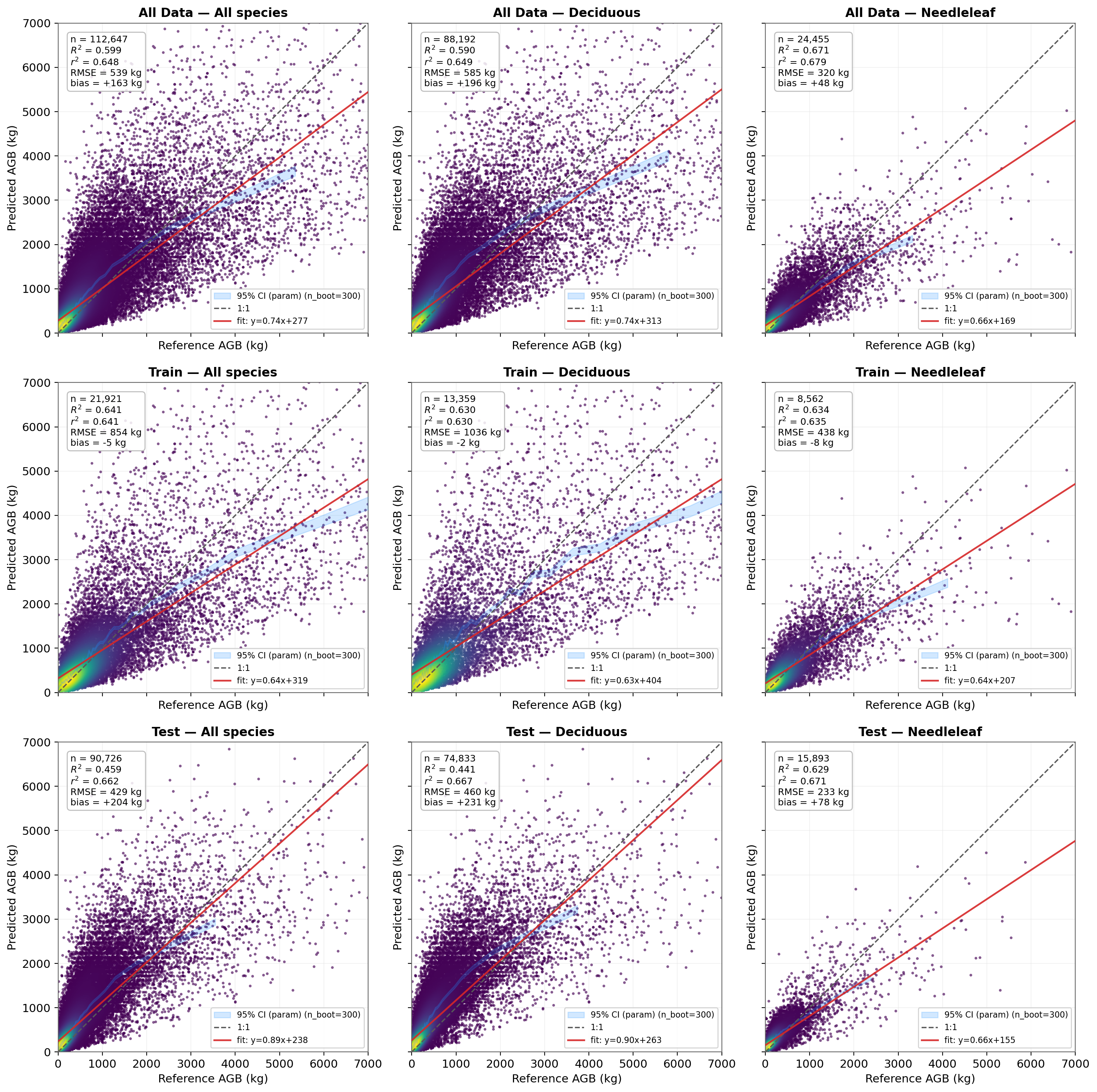}
	\caption{Proxy model fit across the full Oakville inventory ($n = 112{,}647$ trees; top row), with the stratified training subset (middle row) and held-out test subset (bottom row). Columns show all species (left; $n_{\text{train}} = 21{,}921$, $n_{\text{test}} = 90{,}726$), deciduous trees (center; $n_{\text{train}} = 13{,}359$, $n_{\text{test}} = 74{,}833$), and needleleaf trees (right; $n_{\text{train}} = 8{,}562$, $n_{\text{test}} = 15{,}893$). The dashed line is the 1:1 reference; the solid red line is the linear fit. Fitted model parameters correspond to Table~\ref{tab:proxy_params}.}
	\label{fig:calibration}
\end{figure}

\subsection{Biomass Estimation Accuracy}
\label{sec:results_biomass}

The calibrated proxy model reproduces inventory AGB across the full size range. Across the full inventory, the proxy model achieved $R^2 = 0.599$ (all species), $R^2 = 0.590$ (deciduous), and $R^2 = 0.671$ (needleleaf); held-out test-set performance was $R^2 = 0.459$, $R^2 = 0.441$, and $R^2 = 0.629$, respectively. The lower held-out $R^2$ does not indicate poorer model behavior on the test set: the predicted--reference correlation is comparable or higher there (all-species Pearson $r^2 = 0.66$ on the test set versus $0.64$ on the calibration set), but the coefficient of determination is depressed by a systematic positive bias ($+204$~kg on the test set versus $-5$~kg on the calibration set). This pattern is a direct consequence of the balanced stratified split: the calibration set is enriched in large trees, so the NLLS fit centers its residuals near zero there, whereas the small-tree-dominated test set is subject to the proxy's tendency to overestimate small crowns and underestimate large ones, yielding a net positive bias \citep{Jucker2017}. The gap therefore reflects the deliberate calibration design and the natural size distribution of the held-out trees rather than a failure to generalize. The proxy fit is illustrated in Fig.~\ref{fig:calibration}.

\begin{figure}[t]
	\centering
	\includegraphics[width=0.95\textwidth]{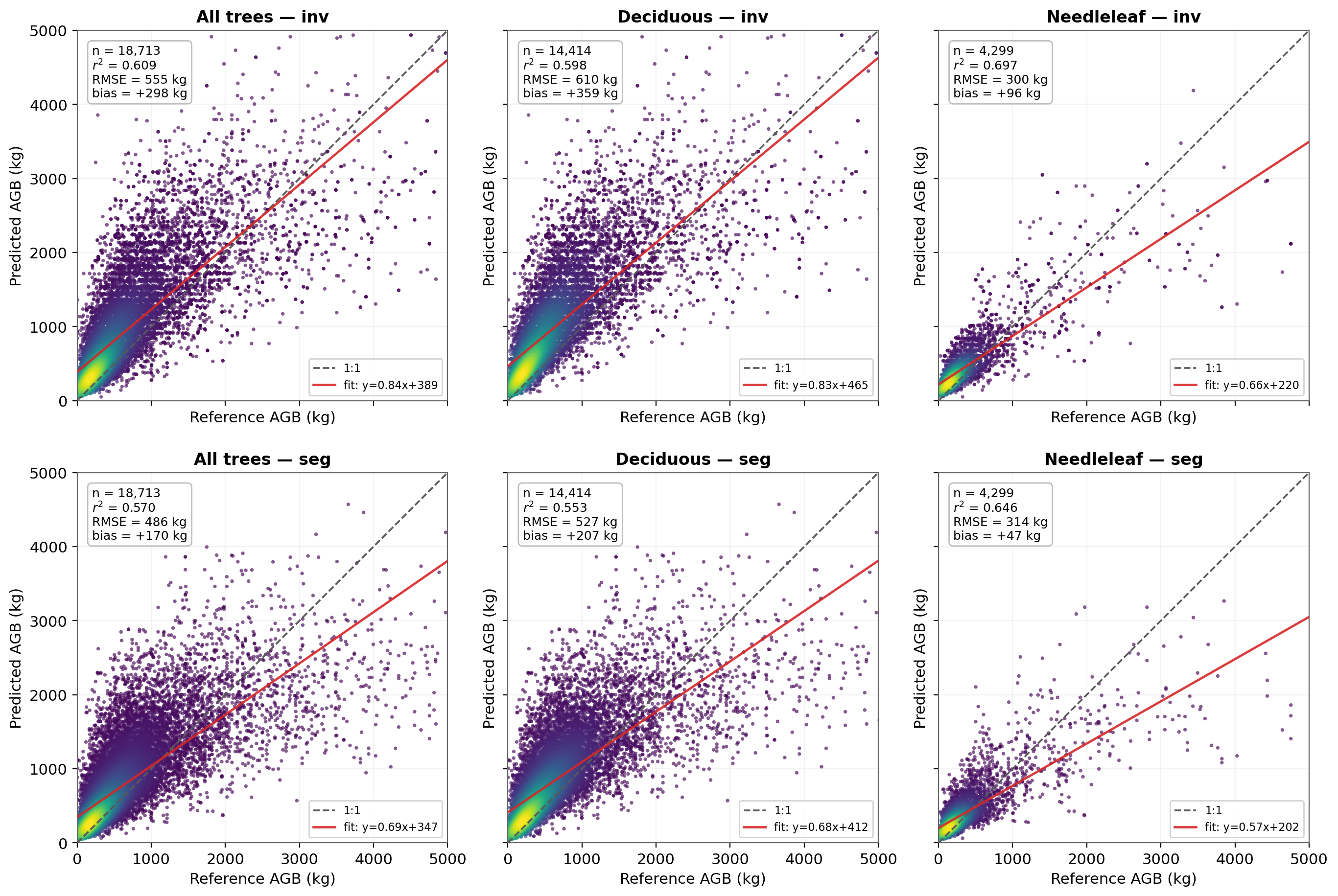}
	\caption{Biomass estimation validation for the \textit{inv} (top row) and \textit{seg} (bottom row) scenarios on the matched validation set ($n = 18{,}713$). Columns show all trees (left), deciduous trees (center), needleleaf trees (right). Each panel reports sample size, $R^2$, RMSE, and bias. Dashed line: 1:1 reference; solid red line: linear fit. Full per-group slopes and intercepts are reported in Supplementary Table~\ref{tab:biomass_supp}.}
	\label{fig:agb_comparison}
\end{figure}

For the matched validation set ($n = 18{,}713$), the \textit{inv} scenario yielded $R^2 = 0.609$, RMSE $= 554.6$~kg, bias $= 298.5$~kg, slope $= 0.84$, and intercept $= 388.6$~kg, whereas the operational \textit{seg} scenario yielded $R^2 = 0.570$, RMSE $= 486.3$~kg, bias $= 170.4$~kg, slope $= 0.69$, and intercept $= 347.2$~kg (Fig.~\ref{fig:agb_comparison}, top vs.\ bottom row). The same pattern was observed for deciduous trees ($n = 14{,}414$; \textit{inv} $R^2 = 0.598$, slope $= 0.83$; \textit{seg} $R^2 = 0.553$, slope $= 0.68$) and needleleaf trees ($n = 4{,}299$; \textit{inv} $R^2 = 0.697$, slope $= 0.66$; \textit{seg} $R^2 = 0.646$, slope $= 0.57$). Needleleaf estimates were consistently more accurate than deciduous estimates.

Full per-group statistics, including slope and intercept, are reported in Supplementary Table~\ref{tab:biomass_supp}.

\subsection{Wall-to-Wall AGB Maps and Spatial Patterns}
\label{sec:results_maps}

\begin{figure}[!t]
	\centering
	\includegraphics[width=0.95\textwidth]{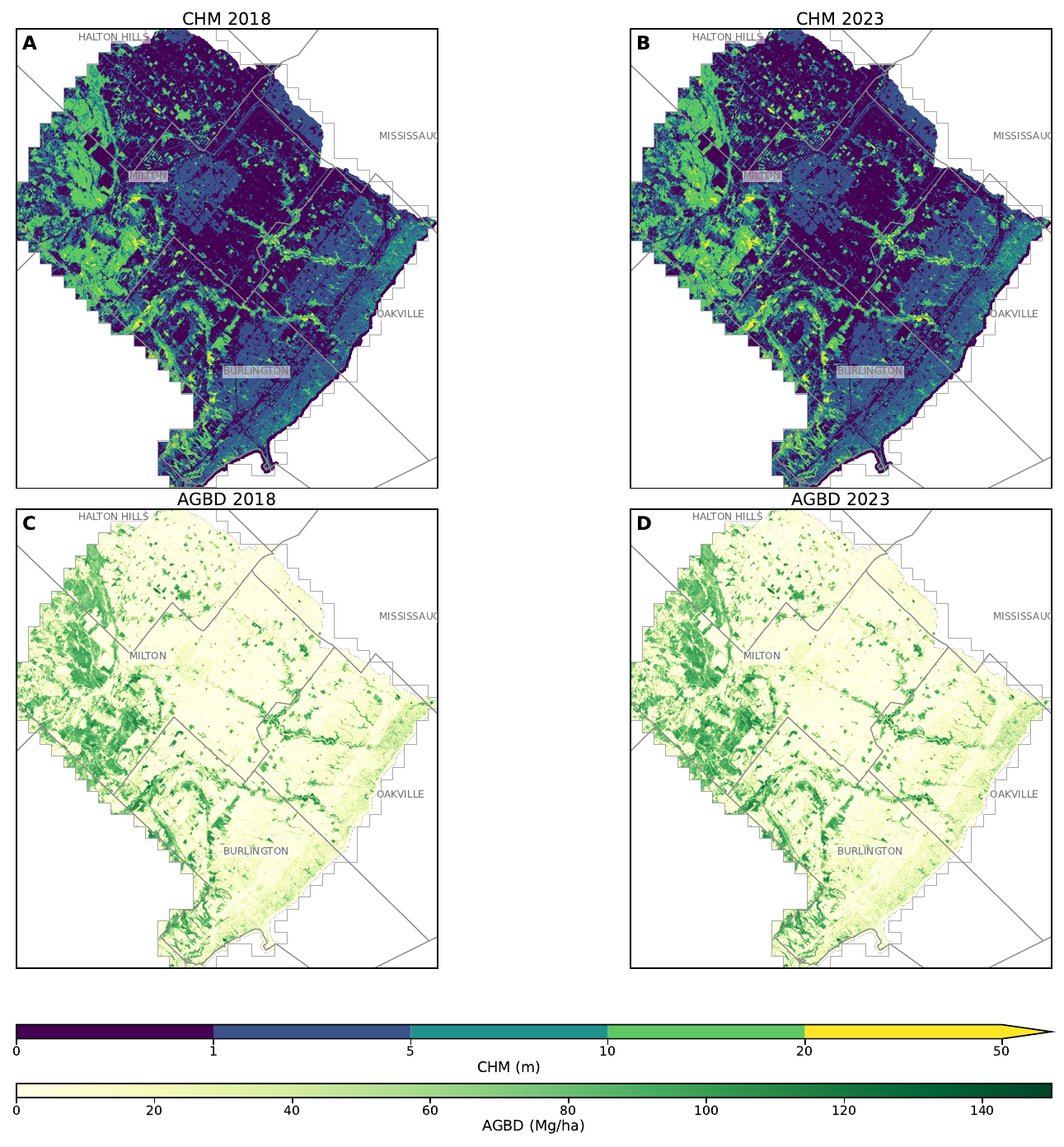}
	\caption{Canopy height models (CHM; top row) and above-ground biomass density (AGBD; bottom row) maps for 2018 (left) and 2023 (right), aggregated to 30~m pixels for study-area-scale visualization and carbon accounting. Crown-level AGB estimates are retained as the native high-resolution product, while the 30~m maps summarize those estimates across the $\sim$810~km$^2$ study area. CHM is displayed using discrete height classes (0--1, 1--5, 5--10, 10--20, and $>$20~m) to improve discrimination of canopy structure, whereas AGBD uses a continuous scale (0--150~Mg~ha$^{-1}$). The 2023 maps extend further northwest than 2018. Municipal boundaries shown for reference.}
	\label{fig:agb_maps}
\end{figure}

The calibrated proxy models were applied wall-to-wall to generate crown-level AGB products for 2018 and 2023. For regional visualization and biomass accounting, these crown-level estimates were aggregated to 30~m raster AGB density (AGBD) maps across the full $\sim$810~km$^2$ study area (Fig.~\ref{fig:agb_maps}). The 30~m grid therefore provides a display and analysis scale rather than the native modeling unit: high-resolution crown-level products are retained for dissemination, but individual-crown maps would be visually unreadable at the full study-area extent in a manuscript figure. Across vegetated pixels (AGBD~$> 0$), the 2018 map covered 51,746~ha with a mean AGBD of 33.34~Mg~ha$^{-1}$ and a total AGB stock of 1,725,288~Mg (carbon stock: 810,885~Mg~C at $f = 0.47$). In 2023, the vegetated area expanded to 52,968~ha, mean AGBD increased to 34.15~Mg~ha$^{-1}$, and total AGB stock rose to 1,808,661~Mg (850,071~Mg~C). Low-biomass pixels (AGBD~$< 10$~Mg~ha$^{-1}$) dominated both maps (69.92\% in 2018; 69.46\% in 2023), reflecting the prevalence of open agricultural land and sparse street-tree canopy, and no pixels exceeded 150~Mg~ha$^{-1}$ in either year. AGBD reached approximately 140~Mg~ha$^{-1}$ along the Niagara Escarpment, in riparian corridors, and in remnant forest patches, whereas residential neighborhoods with sparse street-tree plantings and commercial districts showed systematically lower densities.

\subsection{Temporal AGB Change, 2018--2023, and Ensemble Uncertainty}
\label{sec:results_change}

\begin{figure}[!t]
	\centering
	\includegraphics[width=0.95\textwidth]{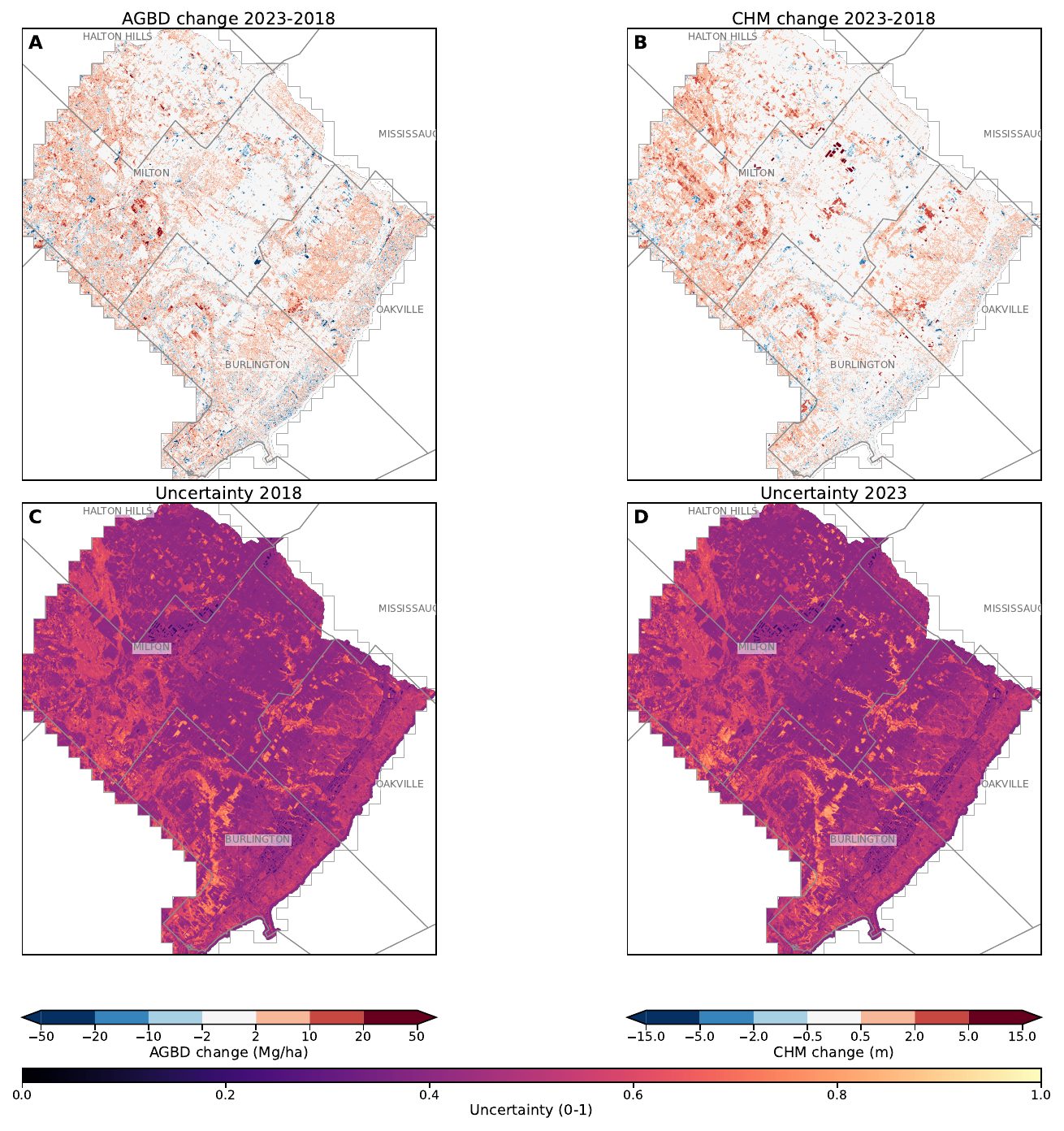}
	\caption{Temporal change and uncertainty maps across the study area at 30~m resolution. (A) AGBD change 2023$-$2018, shown on a discrete diverging scale with class breaks at $\pm$2, $\pm$10, $\pm$20, and $\pm$50~Mg~ha$^{-1}$ (blue tones indicate biomass loss, red tones indicate biomass gain, and the central class denotes negligible change). (B) CHM change 2023$-$2018, with class breaks at $\pm$0.5, $\pm$2, $\pm$5, and $\pm$15~m (same color convention). (C, D) Pixel-level ensemble uncertainty (0--1 scale; purple low, yellow high) for 2018 and 2023, respectively. Municipal boundaries shown for reference.}
	\label{fig:agb_change}
\end{figure}

Fig.~\ref{fig:agb_change} shows the spatial distribution of AGBD change and CHM change between 2018 and 2023, alongside pixel-level uncertainty maps from the deep ensemble. The temporal change analysis was performed over a 77,342~ha valid overlap footprint, defined as all pixels with valid data in both epochs (including non-vegetated cells); this footprint exceeds the per-year vegetated area of $\sim$52{,}000~ha because it includes zero-AGB agricultural and impervious pixels. Net AGB change was $+$83,373~Mg ($+$39,185~Mg~C), with a gross gain of 158,588~Mg and a gross loss of 75,216~Mg. Gain affected 50.38\% of pixels at a mean rate of 0.814~Mg~ha$^{-1}$~yr$^{-1}$, loss affected 20.18\% at a mean rate of $-$0.964~Mg~ha$^{-1}$~yr$^{-1}$, and 29.45\% of pixels showed no detectable change. The CHM change map (Fig.~\ref{fig:agb_change}B) confirms that the localized disturbances visible in the AGBD difference correspond to genuine structural changes in canopy height rather than to artifacts of radiometric or allometric differences between years.

Pixel-level ensemble uncertainty (Fig.~\ref{fig:agb_change}C--D) is lowest in dense forest patches and urban-core areas, where the self-supervised training signal is strongest (well-defined LiDAR planarity for buildings, clear NDVI contrast for vegetation), and highest in agricultural and transitional zones, where the pseudo-labels provide a less discriminative signal; the spatial pattern is consistent across both years. Of all delineated crown segments, 13.50\% (759,185 of 5,623,901) in 2018 and 19.86\% (1,003,724 of 5,053,934) in 2023 were flagged as uncertain ($f_{\text{uncertain}} \geq 0.5$) and routed to the pooled all-trees allometric equation. The higher uncertain fraction in 2023 is consistent with the broader spatial coverage of that acquisition epoch, which extends into more agricultural and transitional zones along the northwest margin of the study area. Across geographic folds, the uncertain fraction ranged from 11.88\% (Set\_2) to 26.02\% (Set\_3), reflecting the uneven distribution of agricultural and mixed land-cover types.

\section{Discussion}

\subsection{Self-Supervised Learning Effectiveness}

The self-supervised multi-modal fusion framework separates vegetation from buildings in complex urban environments without manually labeled training data. By generating pseudo-labels from rule-based heuristics (LiDAR planarity analysis and spectral indices) and refining them through a dual-stream neural network with cross-attention fusion, the framework attained consistent improvements in segmentation accuracy over the pseudo-labels themselves (Table~\ref{tab:segmentation}). The model is trained on the pseudo-labels and evaluated against independently annotated tiles, and its predictions agree more closely with the manual annotations than the pseudo-labels do. The largest gains were observed for buildings (recall $+9.5\%$ and IoU $+11.0\%$ over the pseudo-labels), confirming that LiDAR planarity and optical texture jointly reduce building--vegetation confusion. Reliable building separation is operationally important, because building--vegetation confusion would otherwise propagate directly into false crown delineations and biased biomass estimates. The multitask reconstruction objective further encouraged spatially coherent features that generalize beyond local pseudo-label noise \citep{He2022, Reed2022}.

The necessity of multi-modal fusion is particularly acute under leaf-off acquisitions. Established RGB-only detectors such as DeepForest \citep{Weinstein2019}, Detectree2 \citep{Ball2023}, and SAM \citep{Kirillov2023} are trained predominantly on leaf-on imagery; in leaf-off deciduous forests, bare branches and brown or gray crowns produce low spectral contrast and yield high omission rates and fragmented crown predictions \citep{Chadwick2020, Boyd2022}. The proposed framework addresses this limitation by fusing LiDAR-derived structural cues with optical spectral indices, ensuring that tree detection relies on geometric signals that persist across phenological stages \citep{Dietenberger2023}.

\subsection{AGB Accuracy in Context and the Role of Crown Delineation}
\label{sec:disc_agb}

Reported $R^2$ values across the urban ITC AGB literature span a wide range that largely tracks operational scope rather than sensor or algorithm choice. Comparable studies operate within enclosed parks, gardens, or experimental plots ranging from 0.14 to 6.35~km$^2$ (Supplementary Table~\ref{tab:literature_supp}): \citet{Liu2024} reported $R^2 = 0.81$--0.93 in the 766~km$^2$ Saihanba plantation, using TLS-derived stems for near-direct calibration; \citet{Figueiredo2024} achieved $R^2 = 0.52$--0.70 in the 1.44~km$^2$ Rio Botanical Garden with UAV hyperspectral and LiDAR; \citet{WangShanghai2025} reported $R^2 = 0.91$--0.96 in the 6.35~km$^2$ Haiwan Park for three target species; and \citet{So2025} reported $R^2 = 0.29$--0.80 across 0.14~km$^2$ of rural plots using a methodologically equivalent self-supervised UAV pipeline. The present \textit{seg} $R^2 = 0.570$--0.646 was obtained at $\sim$810~km$^2$ on operational provincial sensor data, without the restricted spatial extent ($<$10~km$^2$), absence of buildings, near-direct (TLS) calibration, or species stratification that characterize the higher-accuracy studies.

The \textit{inv} vs.\ \textit{seg} comparison provides direct evidence that crown delineation is the primary bottleneck in the AGB pipeline. The transition from \textit{inv} to \textit{seg} reduced $R^2$ from 0.609 to 0.570 and the regression slope from 0.84 to 0.69 for all trees; the slope reduction is particularly informative because it indicates that segmentation errors disproportionately affect large trees, which are more likely to be under-segmented owing to irregular crown shapes, overlapping neighbors, and the limitations of watershed segmentation in multi-layered canopies. This effect is amplified by the fact that the largest 10\% of trees account for 56.0\% of total inventory biomass, so large-tree segmentation errors propagate to landscape-scale AGB uncertainty far more than their numerical frequency would suggest. The smaller \textit{seg} bias (170.4~kg versus 298.5~kg for all trees) reflects a partial compensation between this under-segmentation of large trees and the proxy's tendency to overestimate small trees, so slope and $R^2$ characterize structural agreement more directly than bias alone. The superior needleleaf performance ($R^2 = 0.646$ \textit{seg} vs.\ 0.553 for deciduous trees) reflects the structural advantages of needleleaf canopies for CHM-based watershed segmentation: retained canopy volume produces distinct, well-defined height peaks, whereas deciduous bare-branch canopies yield noisier, lower-amplitude CHM profiles. Recent advances in deep learning-based instance segmentation (Mask R-CNN \citep{He2017MaskRCNN}, DETR \citep{Carion2020}, and foundation models for remote sensing \citep{Weber2024}), combined with multi-modal inputs, offer promising avenues to address overlapping crowns and irregular shapes \citep{Dersch2023, Weinstein2019}.

\subsection{Leaf-Off Acquisition Timing}
\label{sec:disc_leafoff}

Leaf-off (spring) acquisition is standard practice in Ontario's provincial mapping programs \citep{OntarioElevationProgram, OntarioImageryProgram} and has important implications for the deciduous-dominated study region (78.3\% deciduous). Leaf-off conditions offer substantial advantages for ground surface modeling: \citet{Simpson2017} reported a DTM RMSE of 0.11~m under leaf-off conditions and 0.23~m in leaf-on deciduous forests, and because $\text{CHM} = \text{DSM} - \text{DTM}$, these errors propagate directly to the estimated canopy heights. For canopy height estimation, high-resolution airborne LiDAR (8--10~pulses~m$^{-2}$) shows relatively small seasonal effects \citep{Chen2022, Cushman2023, Rajab2023}, and the proxy model used here relies on top-of-canopy metrics (maximum crown height and crown area) that are robust to seasonal variation, rather than on within-canopy percentiles. For crown delineation, leaf-off conditions expose trunk and lower-canopy structure: watershed-based methods drop from 87.8\% accuracy in sparse stands to 65.5\% in dense leaf-on canopy \citep{Berra2020}, and the primary failure modes (over-segmentation from within-crown texture and under-segmentation of fused crowns) are reduced under leaf-off conditions, where height-based markers are more distinct \citep{Yang2017, Nuijten2019, Vandendaele2021}.

\subsection{Implications for Regional Carbon Accounting}
\label{sec:disc_carbon}

The crown-level AGB products address an important gap in regional carbon accounting by providing spatially explicit biomass estimates at resolutions actionable for urban forest management. Official carbon accounting frameworks, such as Canada's National Inventory Report \citep{ECCC2026}, rely on inventory- and sample-based methods to estimate forest carbon stocks and stock changes. These approaches provide robust aggregate estimates but are not designed to resolve individual tree crowns or the fine-scale spatial heterogeneity of urban and peri-urban tree cover. Similarly, carbon mapping approaches based on coarse-to-moderate resolution remote sensing products, such as 30~m Landsat or 10~m Sentinel-2, cannot consistently capture individual crowns in heterogeneous urban forests \citep{Song2022, Liang2025}. The workflow therefore separates product resolution from reporting resolution: AGB is estimated and delivered at the crown level, whereas Fig.~\ref{fig:agb_maps}, the study-area totals, and the temporal change summaries use 30~m aggregation for regional display and accounting across the 810~km$^2$ landscape. Across the study area, total AGB stock was 1,725,288~Mg (810,885~Mg~C) in 2018 and 1,808,661~Mg (850,071~Mg~C) in 2023, with mean carbon densities of 15.67 and 16.05~Mg~C~ha$^{-1}$ and a net carbon sequestration (associated with AGB) of $+$39,185~Mg~C over the period. These estimates are conservative: the calibration and validation inventory covers only publicly managed trees in Oakville (street trees, parks, and municipal properties), and the proxy models were applied wall-to-wall, including in municipalities without equivalent inventory data, so a transferability bias may exist if species composition or growing conditions differ systematically. Trees on private residential and commercial lands are excluded from both calibration and validation. Spatial patterns reveal that high-biomass areas are concentrated in vegetated corridors, riparian buffers, and remnant forest patches along the Niagara Escarpment, whereas approximately 70\% of vegetated pixels with low AGBD ($<$50~Mg~ha$^{-1}$) correspond to residential neighborhoods with sparse street-tree plantings, commercial districts, and agricultural land. The crown-level products can therefore inform equitable tree-planting strategies that prioritize underserved neighborhoods. The deep-ensemble uncertainty maps further indicate where these products are least reliable: confidence is lowest in agricultural and transitional zones underrepresented during calibration, and because this spatial pattern is consistent across both acquisition years it reflects systematic data-coverage limitations rather than stochastic noise, flagging where field validation should be prioritized.

The net carbon gain observed over the five-year period is consistent with biomass accumulation through growth in established stands and canopy expansion from new plantings, partly offset by localized losses. The present analysis does not include per-event records of the drivers of change, so these associations are interpretive rather than directly documented; nevertheless, the spatial pattern and magnitude of the losses are consistent with routine urban-forest management (pruning and tree removal), land-use conversion, and ongoing ash mortality from the emerald ash borer (\textit{Agrilus planipennis}), which has devastated green ash populations across southern Ontario since the early 2000s \citep{herms2014emerald}---green ash being among the most abundant species in the study-area inventory. Attributing individual change pixels to specific causes would require integrating municipal management and planting records with pest-monitoring data, and is identified as a direction for future work.

\subsection{Limitations and Future Directions}
\label{sec:disc_limitations}

Some limitations warrant discussion. (1) The manually annotated tile set used for independent segmentation evaluation is necessarily limited in size and may not represent all landscape types, rare tree species, or complex multi-layered canopies. (2) Of the 90,726 held-out test trees, only 18,713 (20.6\%) yielded valid matched pairs with production segments after quality-control filtering (height ratio, density, removal status); the \textit{seg} metrics therefore characterize accuracy on confidently matched crowns, and trees affected by growth between the inventory and LiDAR dates or by under-segmentation are under-represented in this subset. (3) Regression slopes below 1.0 in both scenarios indicate that the power-law functional form does not fully capture the nonlinear scaling of biomass with tree size, inducing a saturation bias for the largest trees, which dominate total AGB \citep{Fischer2019, Aguilar2019, Jucker2017}. Crown-shape predictors (depth, asymmetry) or local competition indices could reduce this bias but require denser point clouds; post-hoc distributional correction such as quantile mapping \citep{Themessl2012, Cannon2015}---widely used in climate modeling but largely unexplored for remote-sensing biomass---is a complementary route, though it is constrained here by the limited, managed-tree reference. (4) The reference AGB used for calibration and validation was derived from the \citet{Lambert2005} national allometric equations, which were fitted to destructively sampled trees from Canadian managed and natural forests. Open-grown urban trees commonly depart from closed-canopy forest architecture---they tend to be shorter for a given stem diameter, more strongly tapered, and to carry proportionally larger crowns---so applying forest-derived DBH-based allometry to municipal street and park trees can introduce a systematic bias in the reference biomass that propagates to the calibrated crown proxy and all downstream products \citep{Aguilar2019, mcpherson2016urban}. Because local destructive-harvest data were not available, the magnitude of this bias---which is shared by urban biomass studies that rely on forest allometry---could not be quantified here. As such a bias is largely multiplicative, the relative spatial and temporal patterns reported here are robust to it, whereas the absolute stocks should be read as a first spatially-explicit estimate; region-specific urban allometry is a priority for refining the absolute magnitudes. (5) All calibration and validation data originate from a single municipal inventory, so multi-site validation is required to assess transferability across urban forest contexts. (6) The temporal change results represent gross raster-level differences; a formal minimum detectable change analysis that propagates classification and allometric uncertainty is a priority for future work. Beyond these, deep learning-based instance segmentation could reduce large-tree crown delineation errors, and finer species identification from hyperspectral imagery or deep learning classification would enable species-specific allometry in place of the broad deciduous/needleleaf functional groups used here.

\section{Conclusions}

This study developed and validated a self-supervised multi-modal fusion framework for high-resolution above-ground biomass estimation in urban environments. By integrating airborne LiDAR (8--10~pulses~m$^{-2}$) and near-infrared RGB orthophotography (0.16--0.20~m) through a dual-stream neural network with cross-attention fusion, the framework achieved accurate vegetation--building separation without manually labeled training data and produced operational crown-level biomass estimates across $\sim$810~km$^2$ of southern Ontario. The model improved on the rule-based pseudo-labels, reaching a mean precision, recall, and Dice coefficient of 0.86, 0.83, and 0.84 against independently annotated tiles, with buildings detected most reliably (Dice 0.95; recall increased 9.5\% over the pseudo-labels). The dual-scenario validation quantified the contribution of crown delineation error to biomass uncertainty: AGB prediction achieved $R^2 = 0.609$ when computed from inventory crown geometry and $R^2 = 0.570$ under operational segmentation-derived geometry, and the slope reduction from 0.84 to 0.69 indicates a systematic underestimation of large-tree biomass attributable to under-segmentation. Needleleaf trees were estimated more accurately ($R^2 = 0.646$ \textit{seg}) than deciduous trees ($R^2 = 0.553$ \textit{seg}), reflecting the structural advantages of retained needleleaf canopy volume for CHM-based watershed segmentation. Aggregation of the crown-level estimates to 30~m AGBD grids for regional reporting yielded total AGB stocks of 1.73~Tg in 2018 and 1.81~Tg in 2023 (810,885 and 850,071~Mg~C, respectively), with densities reaching $\sim$140~Mg~ha$^{-1}$ along the Niagara Escarpment, and a net carbon gain of 39,185~Mg~C over the five-year period. Pixel-level uncertainty maps from the deep ensemble identified regions of high epistemic uncertainty in agricultural and transitional zones underrepresented in the training distribution. The framework requires no manually labeled training data, operates exclusively on standard provincial remote sensing products, and provides a scalable pathway to overcome the systematic underestimation of urban tree biomass in current national carbon inventories---particularly for trees outside forests, which have historically been excluded from forest carbon budget reporting.

\section*{Acknowledgments}
This research was supported by Environment and Climate Change Canada (ECCC). Computational resources were provided by the Digital Research Alliance of Canada (formerly Compute Canada Alliance), enabling the distributed training of the 25-model deep ensemble on high-performance computing infrastructure. We thank the Town of Oakville for providing the municipal tree inventory data, and the Ontario Ministry of Natural Resources for coordinating the High-Resolution Orthophotography Program and Elevation Mapping Program. The findings and views described herein do not necessarily reflect those of Planet Labs PBC.

\bibliographystyle{apalike}
\bibliography{references}

\clearpage
\setcounter{section}{0}
\renewcommand{\thesection}{S\arabic{section}}
\renewcommand{\thetable}{S\arabic{table}}
\renewcommand{\thefigure}{S\arabic{figure}}
\renewcommand{\theequation}{S\arabic{equation}}
\renewcommand{\theHsection}{supp.\arabic{section}}
\renewcommand{\theHtable}{supp.\arabic{table}}
\renewcommand{\theHfigure}{supp.\arabic{figure}}
\renewcommand{\theHequation}{supp.\arabic{equation}}
\setcounter{table}{0}
\setcounter{figure}{0}
\setcounter{equation}{0}

\section*{Supplementary Material}

\section{Supplementary Methods}
\label{sec:supp_methods}

\subsection{Empirical Multi-Sensor Co-registration Assessment}
\label{sec:supp_coregistration}

To characterize the actual distribution of residual spatial offsets between the optical and LiDAR datasets, a translational offset was estimated for each tile by maximizing structural agreement between CHM and optical edge responses. For each valid tile, only structurally informative pixels with canopy height $> 2$~m were retained. The optical image was resampled to the LiDAR grid and converted into a composite edge map by combining Sobel gradient magnitudes from NDVI, grayscale RGB intensity, and the NIR band; a corresponding edge map was extracted from the CHM. Integer translations were evaluated within a bounded search window of $\pm 8$~pixels, and for each candidate shift a normalized cross-correlation (NCC) score was computed between the CHM and shifted optical edge maps. The translation yielding the highest score was taken as the estimated offset. A tile was flagged as potentially misregistered when (i) the best shift exceeded 2~pixels in magnitude, and (ii) the NCC improvement over the zero-shift baseline exceeded 0.03.

Applied to 5,356 valid tiles spanning the full preprocessing footprint across both acquisition years, this procedure flagged 661 tiles (12.3\%) as likely affected by residual misregistration. Across all valid tiles, the median estimated shift was 2.0~pixels (1.00~m at the 0.5~m LiDAR grid resolution); among flagged tiles the median rose to 3.61~pixels (1.80~m), with 90th and 95th percentiles of 8.06 and 8.60~pixels (4.03 and 4.30~m).

These findings motivated a training augmentation strategy: a random spatial translation spanning the empirically observed 95th-percentile offset ($\sim$4.3~m) is applied to the NirRGB input during training (Section~\ref{sec:supp_training}), forcing the network to learn features invariant to the full range of optical--LiDAR misregistration present in the dataset.

\subsection{Radiometric Normalization of 2023 Imagery}
\label{sec:supp_radnorm}

To ensure consistent threshold-based classification across the 2018 and 2023 acquisitions, a bandwise Z-score matching transform was applied to normalize the 2023 optical imagery to the radiometric distribution of the 2018 reference. A visual-diversity-driven tile selection procedure was applied prior to statistics computation to avoid anchoring the transform to the dominant land-cover type (typically bare soil in spring leaf-off imagery), ensuring the reference sample captures the full spectral range of vegetation and built-up classes.

\textbf{Step 1: Candidate tile identification.} Only tiles with valid data for both acquisition years at the same canonical tile location were eligible, ensuring the normalization transfer is meaningful.

\textbf{Step 2: Visual embedding of 2018 tiles.} Each 2018 tile was resized to $224 \times 224$~pixels and passed through a pretrained \texttt{mobilenetv3\_small\_100} backbone \citep{Howard2019} to extract a compact visual feature vector. These embeddings capture appearance similarity in a semantically meaningful feature space, enabling tiles to be grouped by land-cover composition rather than by raw pixel values.

\textbf{Step 3: $k$-means clustering and representative selection.} The embedding vectors were clustered using $k$-means with:
\begin{equation}
k = \max\!\left(2,\; \min\!\left(n,\; \left\lfloor n \cdot r \right\rceil\right)\right), \quad r = 0.3
\label{eq:kmeans_k}
\end{equation}
where $n$ is the number of intersecting tiles. With $r = 0.3$, approximately 30\% of tiles are selected as cluster representatives. From each cluster, the tile closest to the centroid in embedding space was selected, yielding a visually diverse subsample spanning the full range of scene appearances in the study area.

\textbf{Step 4: Z-score matching.} Per-band mean $\mu(b)$ and standard deviation $\sigma(b)$ were computed from the $k$ representative 2018 tiles using Welford's online algorithm \citep{welford1962note}. For each band $b \in \{$Red, Green, Blue, NIR$\}$:
\begin{equation}
I'_{23}(b) = \text{clip}\!\left( \frac{I_{23}(b) - \mu_{23}(b)}{\sigma_{23}(b)} \cdot \sigma_{18}(b) + \mu_{18}(b),\; 0,\; 255 \right)
\label{eq:zscore}
\end{equation}
This normalization was applied \emph{only} during pseudo-label generation; the original unnormalized imagery was retained for neural network training, where natural inter-year radiometric variability acts as implicit augmentation.

\begin{figure}[H]
\centering
\includegraphics[width=\textwidth]{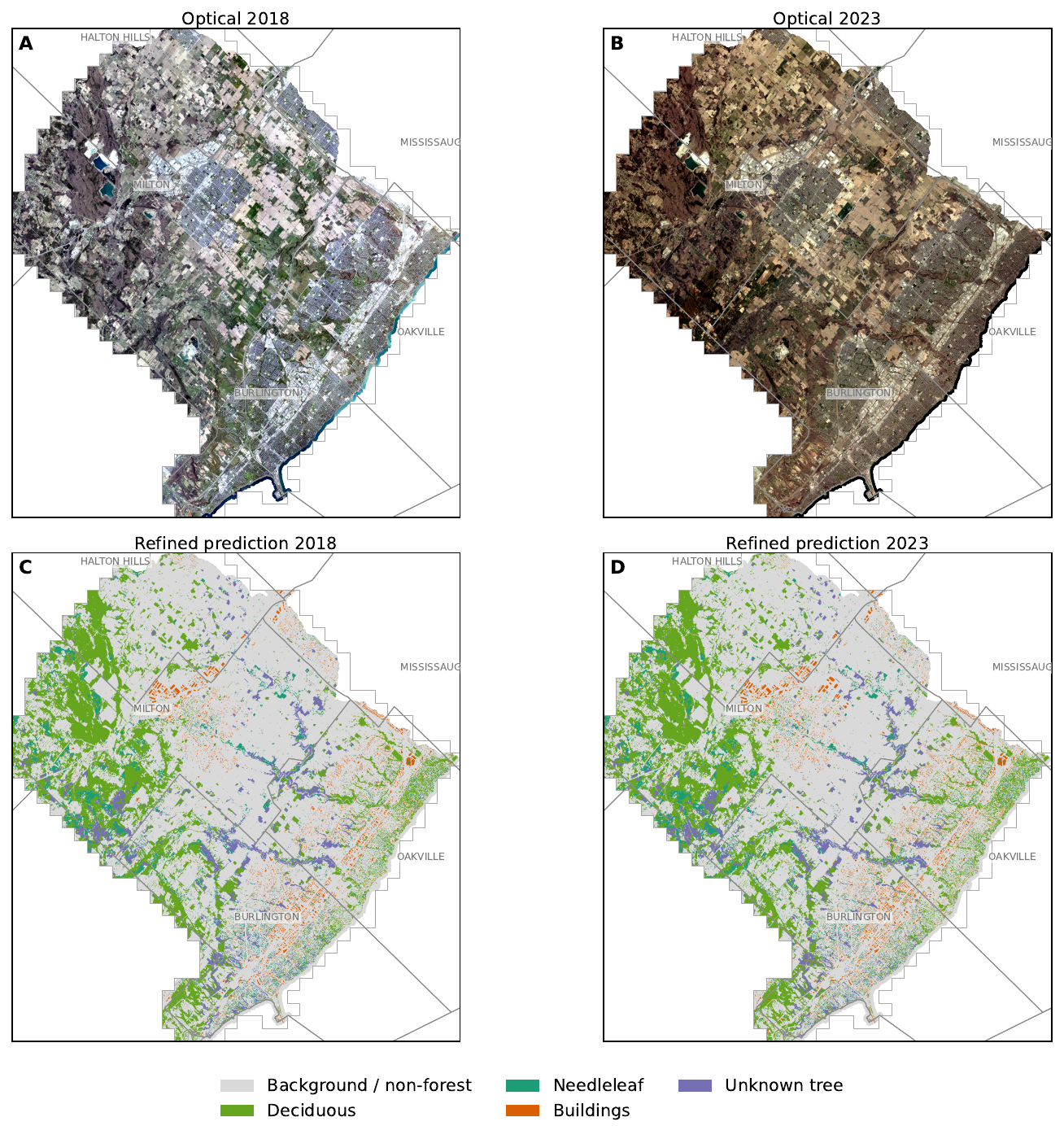}
\caption{Optical imagery and refined land-cover predictions across the $\sim$810~km$^2$ study area. (A, B) NirRGB false-color composites for 2018 and 2023, respectively, displayed as RGB. The pronounced radiometric difference between acquisition years (cool gray tones in 2018 vs.\ warm brown tones in 2023) demonstrates the necessity of bandwise Z-score normalization (Section~\ref{sec:supp_radnorm}) for consistent pseudo-label generation. (C, D) Refined neural network predictions for 2018 and 2023, showing five classes: background/non-forest (gray), deciduous (green), needleleaf (teal), buildings (orange), and unknown tree (purple). All products aggregated to 30~m resolution.}
\label{fig:supp_inputs}
\end{figure}

Fig.~\ref{fig:supp_inputs} illustrates the necessity of this normalization by showing the optical imagery and corresponding refined land-cover predictions for both acquisition years across the full study area. The pronounced radiometric difference between the 2018 (cool blue-gray tones) and 2023 (warm brown-green tones) acquisitions is clearly visible, confirming that without normalization, threshold-based spectral indices (NDVI, HSV criteria) calibrated on 2018 imagery would produce systematically different pseudo-labels when applied to 2023 data.

\subsection{Training Configuration and Experiment Variants}
\label{sec:supp_training}

All 25 ensemble models were trained with the following configuration:

\begin{itemize}[noitemsep]
    \item \textbf{Optimizer:} AdamW \citep{Loshchilov2019}; initial learning rate $2 \times 10^{-4}$; weight decay $0.05$.
    \item \textbf{Learning-rate schedule:} cosine decay with 5-epoch linear warm-up; minimum rate $10^{-6}$.
    \item \textbf{Total epochs:} 75; validation and checkpoint frequency every 5~epochs.
    \item \textbf{Batch size:} 24 (distributed training).
    \item \textbf{Input patch size:} $256 \times 256$~pixels.
    \item \textbf{Data augmentation:} random horizontal/vertical flips, $90^\circ$ rotations, brightness/contrast jitter, and additive CHM noise ($\delta \sim \mathcal{U}[-0.05,\,0.05]$).
    \item \textbf{Co-registration robustness:} random spatial translations ($\Delta x, \Delta y \sim \mathcal{U}[-8,+8]$~pixels) applied to the NirRGB input, spanning the empirically observed 95th-percentile misregistration offset (Section~\ref{sec:supp_coregistration}).
    \item \textbf{Label smoothing:} uniform noise (amplitude~$= 0.2$) added to pseudo-labels during training to prevent overfitting to pseudo-label artifacts \citep{Szegedy2016}.
\end{itemize}

Five experiment configurations per fold were defined to explore sensitivity to key training hyperparameters (Table~\ref{tab:experiments_supp}). The geographic hold-out design prevents spatial leakage between training and validation sets \citep{Roberts2017}.

\begin{table}[H]
\centering
\caption{Experiment configurations per spatial fold (25 total: 5 folds $\times$ 5 configurations). All configurations use the ANL-CE land-cover and tree-type loss with active-term weight $\alpha = 1.0$, softmax temperature $T = 1.0$, masked reconstruction with mask ratio 0.6, $\lambda_{\text{recon}} = 1.5$, and $\lambda_{\text{mask}} = 4.0$; they differ in the ANL-CE passive-term weight $\beta$, the tree-type loss weight $\lambda_{\text{type}}$, and the CHM-confidence weight.}
\label{tab:experiments_supp}
\small
\setlength{\tabcolsep}{4pt}
\begin{tabularx}{\textwidth}{>{\raggedright\arraybackslash}p{1.4cm}>{\centering\arraybackslash}p{1.8cm}>{\centering\arraybackslash}p{2.2cm}>{\centering\arraybackslash}p{2.0cm}>{\raggedright\arraybackslash}X}
\toprule
\textbf{Exp.} & \textbf{ANL-CE $\boldsymbol{\beta}$} & \textbf{Tree-type $\boldsymbol{\lambda_{\text{type}}}$} & \textbf{CHM conf.\ weight} & \textbf{Role} \\
\midrule
\textit{exp\_6}  & 0.5  & 1.0 & 0.4 & Baseline ANL-CE \\
\textit{exp\_8}  & 1.0  & 1.0 & 0.4 & Stronger passive term \\
\textit{exp\_9}  & 0.25 & 1.0 & 0.4 & Weaker passive term \\
\textit{exp\_10} & 0.5  & 1.2 & 0.4 & Stronger tree-type branch \\
\textit{exp\_11} & 0.5  & 1.0 & 0.5 & Stronger CHM weighting \\
\bottomrule
\end{tabularx}
\end{table}

\subsection{Loss Function Equations}
\label{sec:supp_loss}

The multitask loss (Equation~\ref{eq:loss_total} in the main text) comprises the following components.

\textbf{Land-cover loss ($\mathcal{L}_{\text{LC}}$).} An Active Negative Loss with cross-entropy as the active term (ANL-CE) \citep{ye2023active}, a noise-robust loss for learning with noisy labels, computed over the 3-class land-cover labels (background, tree, building):
\begin{equation}
\mathcal{L}_{\text{ANL-CE}} = \alpha\,\mathcal{L}_{\text{NCE}} + \beta\,\mathcal{L}_{\text{pass}}
\end{equation}
where $\mathcal{L}_{\text{NCE}}$ is the normalized cross-entropy (the active term, which drives learning on confidently labeled pixels) and $\mathcal{L}_{\text{pass}}$ is the normalized negative loss (the passive term, which limits overfitting to noisy labels); $\alpha$ and $\beta$ weight the two terms. We fix $\alpha = 1.0$ and vary $\beta \in \{0.25, 0.5, 1.0\}$ across the ensemble (Table~\ref{tab:experiments_supp}). Class-volume-based pixel weights \citep{Diakogiannis2020} handle the imbalance between the large background class and the smaller target classes. The same ANL-CE form, applied only on tree pixels to the 2-class tree-type labels (deciduous, needleleaf), serves as the tree-type loss ($\mathcal{L}_{\text{Type}}$).

\textbf{Reconstruction loss ($\mathcal{L}_{\text{Recon}}$).} A weighted MSE loss applied to both optical and LiDAR reconstruction outputs, with a Laplacian edge-preservation term and emphasis on masked regions:
\begin{equation}
\mathcal{L}_{\text{Recon}} = \lambda_{\text{recon}}\,\text{MSE}_{\text{unmasked}} + \lambda_{\text{mask}}\,\text{MSE}_{\text{masked}} + \lambda_{\text{lap}}\,\mathcal{L}_{\text{Laplacian}}
\end{equation}
with $\lambda_{\text{recon}} = 1.5$, $\lambda_{\text{mask}} = 4.0$, $\lambda_{\text{lap}} = 0.2$. Masking follows a block/grid strategy with mask ratio 0.6 and block size 16~pixels. The higher weight on masked regions forces the encoder to learn representations that predict occluded content from context, encouraging globally coherent features beyond local pseudo-label noise \citep{He2022}. The Laplacian term discourages blurred reconstructions at crown boundaries.

\subsection{Post-Decoder Refinement Modules}
\label{sec:supp_refinement}

After decoding, a set of post-decoder refinement modules progressively sharpens the segmentation representation. First, optical and LiDAR reconstruction logits are predicted from a shared feature space and used to build a \textit{reconstruction-guided gating signal} that modulates the segmentation features---regions where the network reconstructs poorly (typically class boundaries and spectrally ambiguous pixels) are down-weighted before the segmentation heads receive them. The gated representation is then refined by three complementary attention branches applied in parallel:

\begin{enumerate}[noitemsep]
    \item \textbf{Spectral (channel) attention:} recalibrates feature channel importance across the full spatial extent.
    \item \textbf{Local spatial attention:} sharpens fine-grained crown boundary details within a local receptive field.
    \item \textbf{Global spatial attention:} operates on a pooled representation to capture long-range contextual dependencies at reduced quadratic cost.
\end{enumerate}

The outputs of these three branches are combined through a \textit{learnable feature-attention fusion module}, whose weights are optimized jointly with the rest of the network.

\subsection{Uncertainty Decomposition}
\label{sec:supp_uncertainty}

The ensemble of $M = 25$ models produces per-pixel predictive distributions $\{p_m(y \mid \mathbf{x})\}_{m=1}^{M}$. Following the information-theoretic decomposition of \citet{Kendall2017} and \citet{Depeweg2018}, total predictive uncertainty is decomposed into two complementary components.

\textbf{Aleatoric uncertainty} (irreducible data ambiguity) is estimated as the entropy of the ensemble-averaged softmax probabilities:
\begin{equation}
\mathcal{U}_{\text{aleatoric}} = \mathcal{H}\!\left[\bar{p}(y \mid \mathbf{x})\right] = -\sum_{c=1}^{C} \bar{p}_c \log \bar{p}_c, \quad \bar{p}_c = \frac{1}{M}\sum_{m=1}^{M} p_{m,c}
\label{eq:aleatoric}
\end{equation}
Aleatoric uncertainty is highest at class boundaries, in crown overlap zones, and at pixels with inherently ambiguous spectral or structural signatures.

\textbf{Epistemic uncertainty} (reducible model uncertainty) is estimated as the mutual information between the class prediction and the model parameters \citep{Lakshminarayanan2017}:
\begin{equation}
\mathcal{U}_{\text{epistemic}} = \mathcal{H}\!\left[\bar{p}(y \mid \mathbf{x})\right] - \frac{1}{M}\sum_{m=1}^{M} \mathcal{H}\!\left[p_m(y \mid \mathbf{x})\right]
\label{eq:epistemic}
\end{equation}
Epistemic uncertainty is high where ensemble members disagree, identifying areas where the pseudo-label quality or training data coverage is insufficient.

\textbf{Combined crown-level uncertainty.} The per-pixel maps are aggregated to the crown segment level. For each crown segment $S$:
\begin{equation}
f_{\text{uncertain}} = \frac{1}{|S|} \sum_{i \in S} \mathbb{1}\!\left[\mathcal{U}_{\text{total},i} \geq \tau_U\right], \quad \mathcal{U}_{\text{total}} = \mathcal{U}_{\text{aleatoric}} + \mathcal{U}_{\text{epistemic}}
\label{eq:crown_uncertainty}
\end{equation}
where $\tau_U = 0.5$ is a threshold on normalized total uncertainty.

\section{Supplementary Tables}
\label{sec:supp_tables}

\subsection{Literature Comparison: Individual-Tree AGB Studies}
\label{sec:supp_literature}

\begin{table}[H]
\centering
\caption{Comparison of individual-tree AGB studies with directly comparable methodological lineage. Spatial extent, sensor platform, validation sample, and reported $R^2$ are reported alongside the operational regime. All extents are normalized to km$^2$.}
\label{tab:literature_supp}
\scriptsize
\setlength{\tabcolsep}{3pt}
\renewcommand{\arraystretch}{1.08}
\begin{tabularx}{\textwidth}{>{\raggedright\arraybackslash}p{1.8cm}>{\raggedright\arraybackslash}X>{\raggedright\arraybackslash}X>{\raggedright\arraybackslash}X>{\raggedright\arraybackslash}X>{\raggedright\arraybackslash}p{1.25cm}}
\toprule
\textbf{Study} & \textbf{Site \& extent} & \textbf{Context} & \textbf{Sensor \& platform} & \textbf{Validation} & \textbf{$R^2$} \\
\midrule
\citet{Liu2024} &
Saihanba plots within $\sim$766~km$^2$ forest matrix, Hebei, China &
Managed coniferous + broadleaf plantation, rural, closed-canopy &
ALS (airborne) + TLS &
Plot-level, TLS-derived stem dimensions &
0.81--0.93 \\
\addlinespace
\citet{Figueiredo2024} &
Rio de Janeiro Botanical Garden; 1.44~km$^2$ &
Enclosed tropical botanical garden &
UAV hyperspectral + UAV LiDAR &
$>$5{,}600 inventoried trees, single date &
0.52--0.70 \\
\addlinespace
\citet{WangShanghai2025} &
Haiwan National Forest Park, Shanghai; 6.35~km$^2$ &
Single planted urban forest park &
UAV multispectral + TLS &
3 species, plot-level, single date &
0.91--0.96 \\
\addlinespace
\citet{So2025} &
Experimental plots, Southern Ontario; $\sim$0.14~km$^2$ &
Rural forest with thinning treatments + unharvested controls &
UAV LiDAR + UAV RGB &
Plot-level, single date &
0.29--0.47 (thinned) / 0.80 (control) \\
\midrule
\textbf{This study} &
\textbf{This study; $\sim$810~km$^2$ across 8 municipalities, southern Ontario} &
\textbf{Urban--suburban--rural gradient with explicit building separation; 2018 + 2023} &
\textbf{Airborne LiDAR (provincial) + NirRGB (provincial)} &
\textbf{$>$112{,}000 inventoried public trees (Oakville), individual-tree matching, multi-temporal} &
\textbf{0.57--0.65} \\
\bottomrule
\end{tabularx}
\end{table}

\subsection{Full Biomass Estimation Performance Table}
\label{sec:supp_biomass_table}

\begin{table}[H]
\centering
\caption{Biomass estimation performance for inventory-based (\textit{inv}) and segmentation-based (\textit{seg}) crown geometry. Metrics computed on the held-out test set. Slope and intercept refer to the linear regression of predicted vs.\ reference AGB.}
\label{tab:biomass_supp}
\scriptsize
\setlength{\tabcolsep}{4pt}
\begin{tabular}{llrrrrrr}
\toprule
\textbf{Group} & \textbf{Scenario} & \textbf{n} & \textbf{$R^2$} & \textbf{RMSE (kg)} & \textbf{Bias (kg)} & \textbf{Slope} & \textbf{Intercept (kg)} \\
\midrule
\multirow{2}{*}{All trees}  & inv & 18,713 & 0.609 & 554.6 & 298.5 & 0.84 & 388.6 \\
                             & seg & 18,713 & 0.570 & 486.3 & 170.4 & 0.69 & 347.2 \\
\midrule
\multirow{2}{*}{Deciduous}   & inv & 14,414 & 0.598 & 610.2 & 358.9 & 0.83 & 465.3 \\
                             & seg & 14,414 & 0.553 & 526.9 & 207.2 & 0.68 & 411.5 \\
\midrule
\multirow{2}{*}{Needleleaf}   & inv &  4,299 & 0.697 & 300.2 &  95.9 & 0.66 & 219.8 \\
                             & seg &  4,299 & 0.646 & 313.9 &  47.3 & 0.57 & 201.7 \\
\bottomrule
\end{tabular}
\end{table}

\section{Supplementary Figures}
\label{sec:supp_figures}

\begin{figure}[H]
\centering
\includegraphics[width=0.85\textwidth]{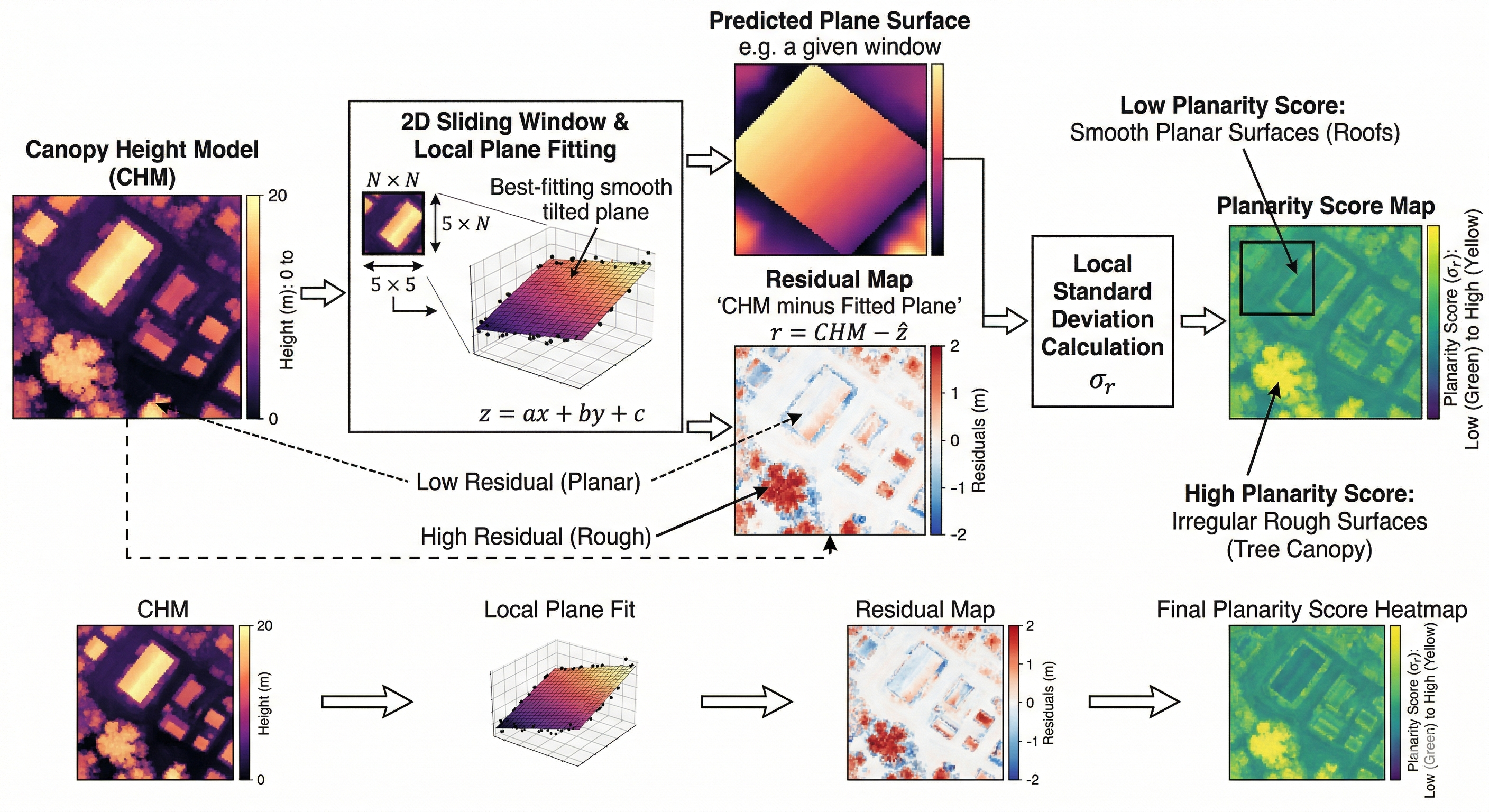}
\caption{LiDAR-based building detection using raster-based plane-fit residual analysis. Building pixels exhibit low residual standard deviation ($\sigma_r < 0.3$~m); vegetation pixels show high values ($\sigma_r > 0.5$~m).}
\label{fig:buildings_lidar_supp}
\end{figure}

\begin{figure}[H]
\centering
\includegraphics[width=0.95\textwidth]{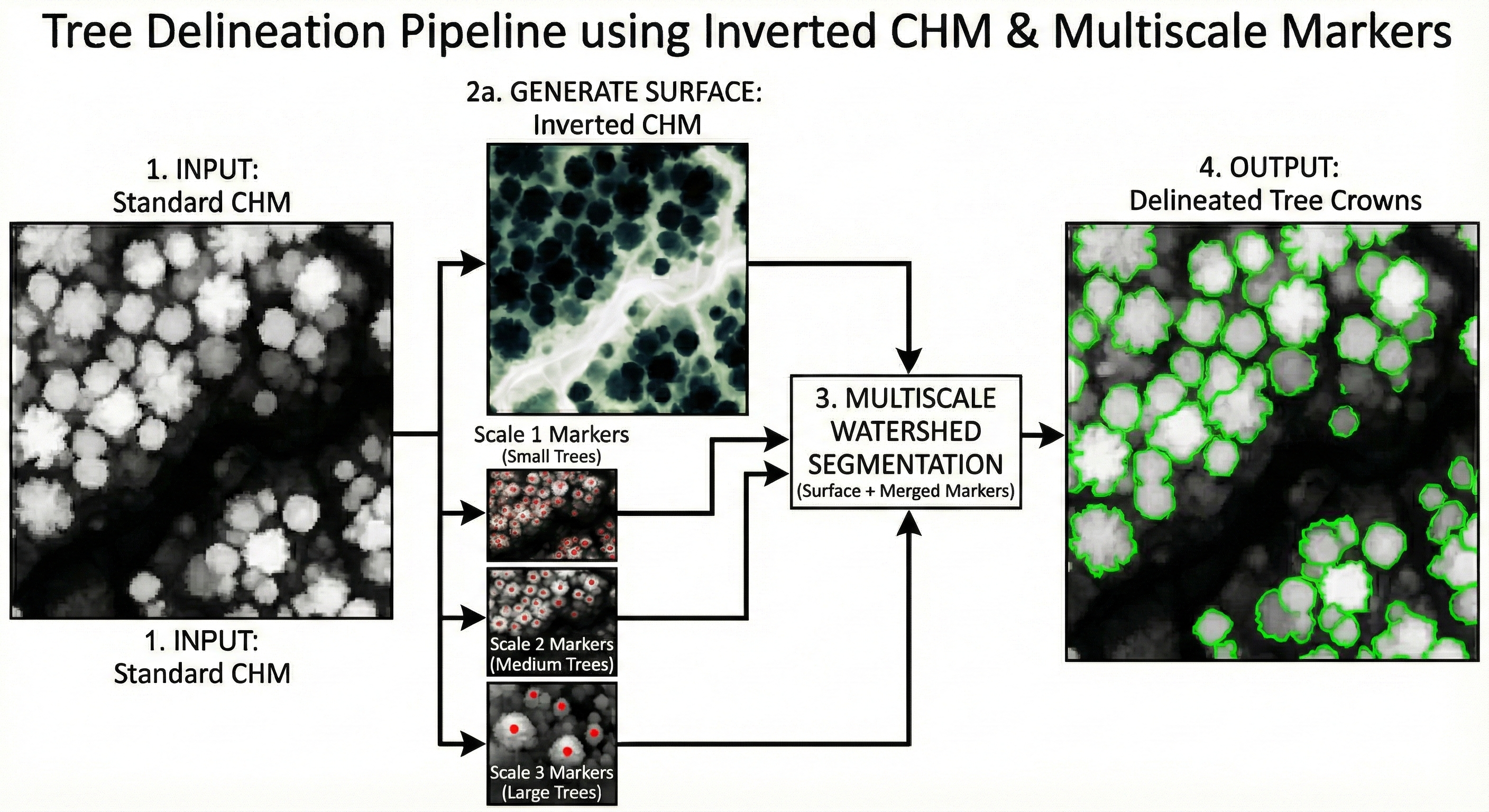}
\caption{Multiscale watershed segmentation for individual tree crown delineation at three spatial scales.}
\label{fig:watershed_supp}
\end{figure}

\begin{figure}[H]
\centering
\includegraphics[width=0.95\textwidth]{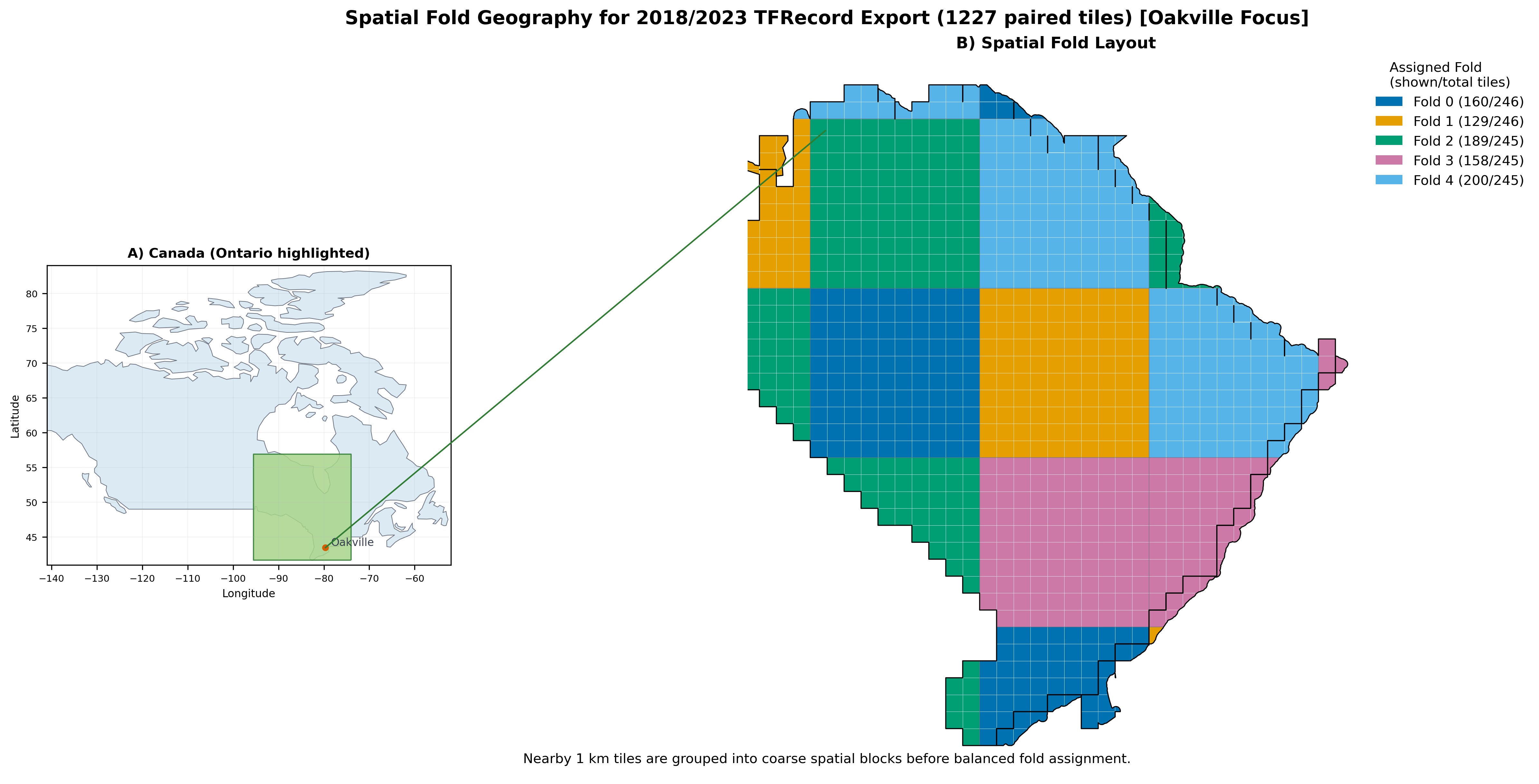}
\caption{Spatial fold layout for geographic cross-validation. The Oakville inventory footprint is partitioned into five non-overlapping geographic regions; at each fold, four regions supply training tiles and the held-out region provides unseen tiles for in-training validation and checkpoint selection.}
\label{fig:geographic_split_supp}
\end{figure}

\begin{figure}[H]
\centering
\includegraphics[width=0.95\textwidth]{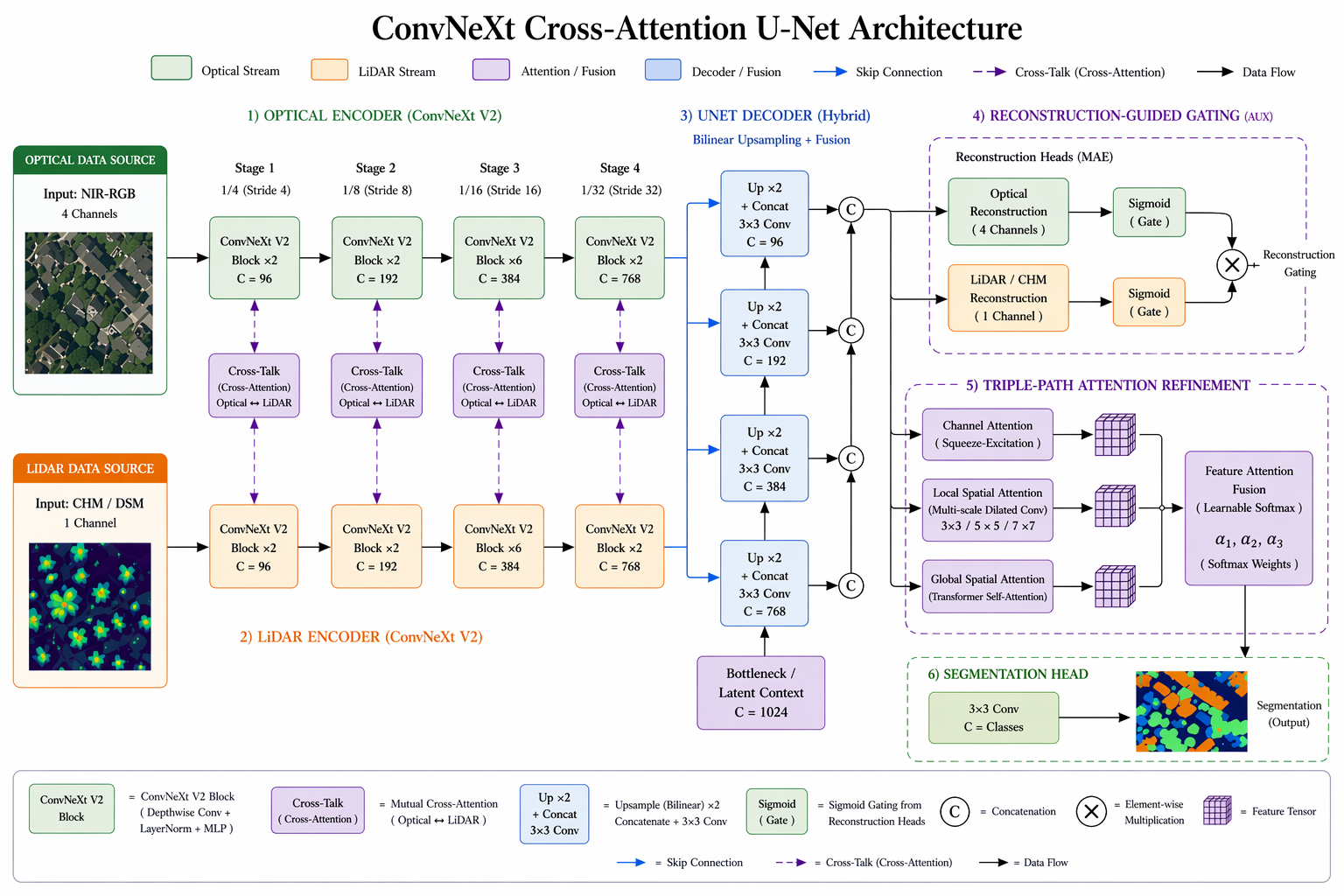}
\caption{Dual-stream cross-attention ConvNeXtV2 U-Net architecture. Optical (NirRGB) and LiDAR (CHM) channels are encoded by modality-specific ConvNeXtV2 backbones; cross-modal attention blocks inserted after each encoder stage exchange structural and spectral features. The U-Net decoder merges both streams via skip connections. Post-decoder refinement modules (spectral, local spatial, and global spatial attention) sharpen the segmentation representation before the land-cover and tree-type prediction heads. Reconstruction heads for both modalities provide masked autoencoder regularization during training.}
\label{fig:nn_training_supp}
\end{figure}

\end{document}